%% file: v5.tex
\documentclass{article} 
\usepackage{iclr2027_conference,times}

\input{math_commands.tex}

\usepackage{hyperref}
\usepackage{url}
\usepackage{booktabs}
\usepackage{graphicx}
\usepackage{multirow}
\usepackage{xcolor}
\usepackage{colortbl}
\usepackage{tabularx}
\usepackage{array}
\usepackage{makecell}
\usepackage{xcolor}
\usepackage{threeparttable}
\usepackage{amsmath}
\usepackage{adjustbox}
\usepackage{placeins}
\usepackage{wrapfig}
\usepackage{float}
\newcommand{\tabcite}[1]{{\footnotesize\citep{#1}}}
\definecolor{resultred}{RGB}{180, 30, 40}

\definecolor{improvementgreen}{RGB}{0, 145, 80}
\definecolor{improvementred}{RGB}{200, 45, 45}

\title{NeurDuo-EEG: A Long-Sequence EEG Foundation Model with Persistent State and Explicit Memory}

\author{%
\parbox[t]{\dimexpr\textwidth-2\tabcolsep\relax}{%
\centering
\setlength{\tabcolsep}{0pt}%
\renewcommand{\arraystretch}{1.25}%
\begin{tabular}[t]{@{}*{4}{>{\centering\arraybackslash\bfseries}p{0.25\linewidth}}@{}}
\mbox{Yifan Wang\textsuperscript{1}\thanks{Equal contribution.}} &
\mbox{Haiping Liu\textsuperscript{1}\footnotemark[1]} &
\mbox{Yang Cui\textsuperscript{1}} &
\mbox{Wenhao Cai\textsuperscript{1}} \\
\mbox{Shuhang Li\textsuperscript{2}} &
\mbox{Xiaoyang Huang\textsuperscript{1}} &
\mbox{Xianyang Liu\textsuperscript{3}} &
\mbox{Jingyu Sun\textsuperscript{1}} \\
\mbox{Yizheng Sun\textsuperscript{1}} &
\mbox{Cunhang Fan\textsuperscript{4}} &
\mbox{Tianming Du\textsuperscript{5,6}} &
\mbox{Jiancheng Yang\textsuperscript{5,6}} \\
\mbox{Zhenhong Li\textsuperscript{1}} &
\mbox{Yunhao Zhang\textsuperscript{7,8}} &
\mbox{Hongpeng Zhou\textsuperscript{1}} &
\mbox{Jingyuan Sun\textsuperscript{1}\thanks{Correspondence to
\href{mailto:jingyuan.sun@manchester.ac.uk}{\texttt{jingyuan.sun@manchester.ac.uk}}.}}
\end{tabular}\\[9pt]
\normalfont\small
\textsuperscript{1}University of Manchester
\qquad
\textsuperscript{2}ETH Z\"{u}rich\\[3pt]
\textsuperscript{3}Shanghai Jiao Tong University
\qquad
\textsuperscript{4}Anhui University\\[3pt]
\textsuperscript{5}ELLIS Institute Finland
\qquad
\textsuperscript{6}Aalto University\\[3pt]
\textsuperscript{7}Institute of Automation, Chinese Academy of Sciences\\[3pt]
\textsuperscript{8}University of Chinese Academy of Sciences\\[6pt]
\texttt{\{yifan.wang, jingyuan.sun\}@manchester.ac.uk}
}
}

\usepackage{wrapfig}

\iclrfinalcopy 
\begin{document}

\maketitle
\fancyhead{}
\renewcommand{\headrulewidth}{0pt}

\begin{abstract}

Electroencephalography (EEG) is recorded continuously over hours, with relevant dynamics spanning timescales from milliseconds to hours. Most EEG foundation models nevertheless process fixed windows independently, limiting their ability to capture information encoded in long-timescale dynamics. State-space architectures enable persistent recurrent processing, but long-range information remains implicitly compressed in recurrent states. We present NeurDuo-EEG, a causal EEG foundation model with channel-resolved persistent memory. NeurDuo-EEG introduces multi-timescale memory management with learned consolidation and selective retrieval, enabling persistent modelling of continuous EEG with fixed-size state. It is pre-trained on 3,955 hours of EEG from 17 public datasets using multichannel autoregressive prediction of discrete spectral codes. Across three short-window and two long-sequence downstream tasks, NeurDuo-EEG achieves the best performance on four of five benchmarks, including all three short-window tasks and seizure detection, where AUC-PR improves from $0.285$ to $0.471$ over the strongest non-NeurDuo baseline. NeurDuo-EEG also remains competitive on sleep staging and supports efficient streaming inference, with nearly constant per-chunk latency as the available history grows to one hour. Notably, the Small variant achieves this with only 4.7M backbone parameters. These results demonstrate the value of persistent, multi-timescale modelling for both long-sequence and short-window EEG analysis. Our code is available at \url{https://github.com/YifaNNW/NeurDuo-EEG}.

\end{abstract}

\section{Introduction}
Electroencephalography (EEG) enables non-invasive monitoring of brain activity over extended periods. 
In applications such as sleep assessment and epilepsy monitoring, recordings often continue for hours or even days \citep{berry2012aasm,tatum2022minimum}. 
Within these long recordings, however, clinically relevant dynamics unfold over very different timescales. 
Epileptiform spikes may last only tens of milliseconds \citep{kane2017revised}, whereas seizure evolution, sleep-stage transitions, and broader changes in brain state develop over minutes to hours. 
At the same time, long recordings are highly redundant, with informative events occurring sparsely amid extended background activity. 
Moreover, EEG is inherently multichannel, with electrodes sampling activity from different scalp locations and potentially exhibiting distinct temporal dynamics.
Modelling continuous EEG therefore requires \textit{short-term sensitivity}, \textit{long-term retention}, \textit{selective access to relevant past information}, and preservation of \textit{channel-specific temporal structure}.

Most existing EEG foundation models follow a window-based paradigm, where self-supervised pre-training is performed on fixed-length segments and each window is subsequently processed independently \citep{jiang2024large,wang2024eegpt,wang2025cbramod,eegreview}. This prevents the backbone itself from carrying temporal state across window boundaries. Extending the window only moves this boundary outward: more history becomes available, but without a mechanism to selectively preserve relevant information, important events may be increasingly buried in redundant context. Moreover, longer windows increase computational and memory costs.

Selective state-space models (SSMs) support linear-time sequence modelling \citep{mamba}, and early Mamba-based EEG models primarily exploited this efficiency within a window-based paradigm \citep{eegmamba,femba}. However, temporal dependencies across successive windows remained unmodelled.
More recently, CaMBRAIN~\citep{CaMBRAIN} took a further step toward continuous EEG modelling by embedding each multichannel EEG patch into a shared token and recurrently modelling the resulting temporal stream with persistent Mamba states.
However, a single recurrent state struggles to preserve both short-term responsiveness and long-term retention. Without explicit retrieval, relevant past representations become difficult to recover once compressed into the evolving state. In addition, fusing multiple channels into a single token may obscure channel-specific temporal dynamics.

To address these challenges, we introduce NeurDuo-EEG (hereafter referred to as NeurDuo), a causal EEG foundation model with channel-resolved, structured persistent memory. NeurDuo maintains fast per-channel recurrent streams to capture short-term dynamics while preserving channel-specific temporal structure, with interleaved cross-channel attention enabling interaction between electrodes. A learned writer periodically consolidates recent fast representations into compact memory tokens, which update a slower recurrent stream for long-term retention and populate a bounded memory bank for selective content-based retrieval.
During pre-training, a frozen spectral tokenizer provides discrete per-channel targets, and NeurDuo is trained autoregressively to predict the next spectral code for all channels in parallel. We evaluate NeurDuo against six EEG foundation models and three classical methods across three short-window and two long-sequence downstream tasks. NeurDuo achieves the best performance on four of five benchmarks, including all three short-window tasks, and improves CHB-MIT seizure-detection AUC-PR from $0.285$ for the strongest non-NeurDuo baseline to $0.471$.
Importantly, NeurDuo maintains nearly constant per-chunk inference latency as the available history grows from $30$s to $1$h, enabling efficient streaming over long recordings.
Overall, NeurDuo establishes a unified framework for modelling continuous multichannel EEG through channel-resolved causal processing and structured memory across temporal scales, while supporting efficient streaming inference over long recordings.




\section{Related Work}
\label{sec:related}
\paragraph{Window-based EEG foundation models.}
Existing EEG foundation models differ in how they tokenize the signal, what they learn during pre-training and how they cope with heterogeneous montages \citep{yang2023biot, wang2025cbramod, wang2024eegpt, yang2026eeg, el2026reve, doner2026luna, zhou2026msbram}, yet they share one design: each takes a bounded input segment and computes its representation from that segment alone. LaBraM, for example, predicts masked vector-quantized codes from context on both sides within the window \citep{jiang2024large}. These backbones encode bounded windows without carrying recurrent state between them. NeurDuo instead reads the recording causally, carrying a fixed-size state across steps, so each new sample has constant cost and context extends beyond window boundaries.

\paragraph{State-space models for EEG.}
State-space models such as Mamba process sequences in linear time through an input-dependent recurrent state, which makes them natural candidates for causal, streaming inference \citep{mamba}. EEG models built on Mamba nonetheless still encode bounded segments, mostly with bidirectional backbones \citep{eegmamba, femba, samba}. CaMBRAIN is the exception, carrying a causal recurrent state across successive EEG patches \citep{CaMBRAIN}, but it merges all channels into one representation before the recurrence and keeps long-range information only implicitly in that state. 
This limits channel-resolved temporal modelling and explicit access to past context.
NeurDuo instead keeps a recurrent stream for each channel and assigns short-term processing, long-term retention and explicit retrieval to a fast state, a slow state and a content-addressable memory, respectively.

\section{Methods}
\label{sec:method}
\begin{figure}[t]
\centering
\includegraphics[width=\linewidth]{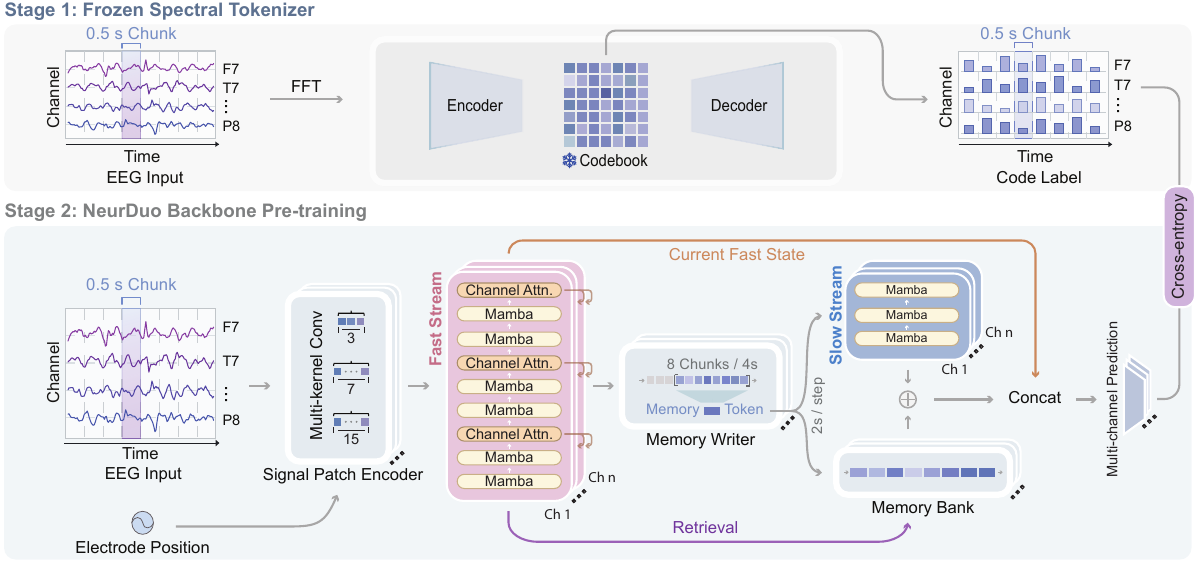}
\caption{\textbf{Overview of NeurDuo and its two-stage pre-training framework.} A frozen spectral tokenizer provides discrete per-channel targets. The backbone processes coordinate-aware EEG streams through fast and slow recurrent pathways, consolidates recent states into bounded memory, and retrieves relevant past information for per-channel next-token prediction.}
\label{fig:framework}
\end{figure}

Pre-training has two stages: we first train and freeze a channel-independent spectral tokenizer, then train NeurDuo to predict the next spectral code for each channel from raw multichannel EEG using a stateful multi-timescale backbone (Figure~\ref{fig:framework}).

\subsection{Spectral tokenizer}
\label{subsec:tokenizer}


All EEG recordings are resampled to $256$\,Hz and divided into non-overlapping $0.5$-s chunks $x_{t,c}\in\mathbb{R}^{L}$, with $L=128$. For each chunk, we compute its one-sided frequency spectrum using the fast Fourier transform (FFT), yielding $F=65$ bins, and standardize its magnitude and phase separately along the frequency axis:
\begin{equation}
\phi_{t,c}=\left[\operatorname{std}\!\left(\left|\mathcal{F}x_{t,c}\right|\right)\;\Vert\;\operatorname{std}\!\left(\angle\mathcal{F}x_{t,c}\right)\right]\in\mathbb{R}^{2F}.
\end{equation}
A two-block pre-norm residual MLP independently maps each channel's spectral representation to $g_{t,c}=E(\phi_{t,c})$, without cross-channel interaction or electrode identity. Encoder outputs and the $V=2048$ codebook vectors are $\ell_2$-normalized, and the discrete code is assigned by cosine similarity, $y_{t,c}=\arg\max_{v\in\{1,\ldots,V\}}\bar g_{t,c}^{\top}\bar e_v$. std denotes frequency-wise z-score normalisation.


The codebook is updated by exponential moving averages. A lightweight decoder reconstructs the standardized magnitude and phase spectra from the selected codes. The tokenizer is trained with
\begin{equation}
\mathcal L_{\mathrm{tok}}=\mathcal L_{\mathrm{amp}}+\mathcal L_{\mathrm{phase}}+\beta\mathcal L_{\mathrm{commit}},\qquad
\mathcal L_{\mathrm{commit}}=\left\|\operatorname{sg}\!\left(\bar e_{y_{t,c}}\right)-\bar g_{t,c}\right\|_2^2.
\end{equation}
$\mathcal L_{\mathrm{amp}}$ and $\mathcal L_{\mathrm{phase}}$ are MSE reconstruction losses, and $\operatorname{sg}$ denotes stop-gradient. The trained tokenizer is then frozen to generate targets $y_{t,c}$ for NeurDuo pre-training.

\subsection{NeurDuo Architecture}
\label{sec:arch}
NeurDuo consists of a coordinate-aware channel encoder that preserves electrode-specific representations, followed by a multi-timescale stateful backbone.

\subsubsection{Coordinate-Aware Channel Encoding}
\label{sec:coordinateAware}
To obtain compact input tokens while preserving channel-specific information, we encode each electrode independently within each temporal chunk and retain the channel axis throughout.

\paragraph{Signal patch encoder.}

Each channel chunk $x_{t,c}$ is independently processed by three parallel two-layer temporal convolutional branches with kernel widths $\{3,7,15\}$, producing multi-scale temporal features. The branch outputs are concatenated and fused by a pointwise convolution to obtain $f_{t,c}\in\mathbb{R}^{H\times L}$, where $H$ is the frontend hidden width. These features are attention-pooled across the $L$ temporal positions using a learned scoring vector $\mathbf{w}\in\mathbb{R}^{H}$ and projected to the fast-stream width $d_f$ by $W_p\in\mathbb{R}^{d_f\times H}$:
\begin{equation}
a_{t,c}=\operatorname{softmax}\!\left(w^\top f_{t,c}\right),\qquad
\bar f_{t,c}=f_{t,c}a_{t,c},\qquad
\tilde u_{t,c}=W_p\bar f_{t,c}.
\end{equation}

\paragraph{Electrode identity.}
To provide each channel-local token with a montage-independent electrode identity, we encode the physical electrode coordinates $p_c\in\mathbb{R}^3$ together with the reference scheme $\rho_c$, using a learned embedding $E_{\mathrm{ref}}$ for the latter. The resulting metadata representation is added to the signal token as
\begin{equation}
e_c=\operatorname{MLP}(p_c)+E_{\mathrm{ref}}[\rho_c]\in\mathbb{R}^{d_f},\qquad
u_{t,c}=\operatorname{LN}\!\left(\tilde u_{t,c}+e_c\right)\in\mathbb{R}^{d_f}.
\end{equation}
At each temporal step, the resulting $C$ channel-specific tokens $\{u_{t,c}\}_{c=1}^{C}$ are retained as parallel streams and serve as inputs to the backbone without channel pooling.

\subsubsection{Multi-Timescale Stateful Backbone}
\label{sec:backbone}
The backbone maintains information at complementary temporal scales. A fast recurrent stream processes channel-specific tokens over time, with interleaved channel attention for cross-channel interaction. A learned writer periodically compresses recent fast representations into memory tokens, which update a slower recurrent stream and populate a bounded bank for content-based retrieval.

\paragraph{Fast stream.}

The fast stream models short-timescale dynamics at each $0.5$-s step using $L_f$ pre-norm Mamba blocks of width $d_f$, applied independently along the temporal dimension of each channel stream. 
Let $h^{(\ell)}_{:,c}$ denote the channel-$c$ sequence after block $\ell$. The temporal update is:
\begin{equation}
    h^{(\ell)}_{:,c}
    =
    h^{(\ell-1)}_{:,c}
    +
    \operatorname{Mamba}^{(f)}_{\ell}
    \left(
    \operatorname{LN}\!\left(h^{(\ell-1)}_{:,c}\right)
    \right),
    \qquad
    h^{(0)}_{t,c}=u_{t,c}.
\end{equation}

To enable cross-channel interaction while preserving channel-specific temporal trajectories, we apply multi-head self-attention across the $C$ channel representations after every second Mamba block. At each fixed temporal step $t$,
\begin{equation}
    h^{(\ell)}_{t,:}
    \leftarrow
    h^{(\ell)}_{t,:}
    +
    \operatorname{MHA}
    \left(
    \operatorname{LN}\!\left(h^{(\ell)}_{t,:}\right)
    \right),
    \qquad
    \ell\equiv0\pmod 2.
\end{equation}

The Mamba blocks model temporal dependencies within each channel, while channel attention integrates information across electrodes at the same temporal step. 
We denote the final contextualized fast representation by $h_{t,c}=h^{(L_f)}_{t,c}\in\mathbb{R}^{d_f}$.

\paragraph{Memory writer.}

To bridge the fast stream with the slow stream and memory bank, the memory writer periodically consolidates recent fast representations into memory tokens.
Every $N=4$ steps ($2$\,s), it summarizes the most recent $w=8$ representations ($4$\,s) of each channel. Let $t_k$ denote the step of the $k$-th write. Using a learnable scoring vector $v$, the writer attention-pools these representations and projects the result to the slow-stream width $d_s$:
\begin{equation}
\alpha_j=\operatorname{softmax}_j\left(v^\top h_{t_k-w+j,c}\right),\qquad
m_{k,c}=W_m\operatorname{LN}\left(\sum_{j=1}^{w}\alpha_j h_{t_k-w+j,c}\right)\in\mathbb{R}^{d_s}.
\end{equation}

\paragraph{Slow stream.}
The memory tokens $m_{k,c}$ form a lower-rate sequence for each channel, which is processed by $L_s$ pre-norm Mamba blocks of width $d_s$ to capture longer-timescale temporal structure. Let $s^{(\ell)}_{:,c}$ denote the channel-$c$ sequence after slow-stream block $\ell$:
\begin{equation}
s^{(\ell)}_{:,c}
=
s^{(\ell-1)}_{:,c}
+
\operatorname{Mamba}^{(s)}_{\ell}
\left(
\operatorname{LN}\!\left(s^{(\ell-1)}_{:,c}\right)
\right),
\qquad
s^{(0)}_{k,c}=m_{k,c}.
\end{equation}
The final slow representation at write $k$ is $s_{k,c}=s^{(L_s)}_{k,c}\in\mathbb{R}^{d_s}$. Since the slow stream advances only when a new memory token is produced, it updates at $1/N$ of the fast-stream rate.

\paragraph{Memory bank.}
In parallel with the slow stream, each memory token $m_{k,c}$ is written to a bounded channel-specific memory bank with capacity $B=8$. New tokens are appended in first-in-first-out order, evicting the oldest entry when full.

\paragraph{Content-addressable retrieval.}
To selectively access past information relevant to the current signal, the fast representation $h_{t,c}$ queries the memory bank of the same channel. Let $b_{j,c}$ denote the $j$-th available memory entry. We obtain the query, key, and value representations as $q_{t,c}=W_qh_{t,c}$, $k_{j,c}=W_kb_{j,c}$, and $v_{j,c}=W_vb_{j,c}$, and compute content-based attention over the bank:
\begin{equation}
    \pi_{t,c,j}
    =
    \operatorname{softmax}_j
    \left(
    \frac{q_{t,c}^{\top}k_{j,c}}{\sqrt{d_s}}
    \right),
    \qquad
    \tilde r_{t,c}
    =
    \sum_j
    \pi_{t,c,j}v_{j,c}.
\end{equation}

A learned gate then controls how much of the retrieved content is incorporated:
\begin{equation}
    r_{t,c}
    =
    \sigma
    \left(
    W_g
    \left[
    h_{t,c}\Vert\tilde r_{t,c}
    \right]
    \right)
    \odot
    \tilde r_{t,c}
    \in\mathbb{R}^{d_s}.
\end{equation}
Attention is restricted to memory entries available at step $t$, preserving causality.

\paragraph{NeurDuo output.}
To combine recurrent long-range context with explicitly retrieved memory, we add the retrieval representation to the current slow representation and concatenate the result with the fast representation. We use $\kappa(t)$ to index the most recent memory write available at step $t$, with $s_{\kappa(t),c}$ denoting the corresponding slow representation. The backbone output is
\begin{equation}
\label{eq:readout}
z_{t,c}
=
\left[
h_{t,c}
\;\Vert\;
\left(
s_{\kappa(t),c}
+
r_{t,c}
\right)
\right]
\in\mathbb{R}^{d_z},
\qquad
d_z=d_f+d_s.
\end{equation}

\subsection{MultiChannel Autoregressive Pre-training}
\label{sec:multichannel}

To preserve multichannel structure and train NeurDuo causally, we use an autoregressive objective that predicts the next spectral code for all channels in parallel. The frozen spectral tokenizer maps each next chunk $x_{t+1,c}$ to its discrete code $y_{t+1,c}\in\{1,\ldots,V\}$, which serves as the prediction target. A prediction head shared across channels maps $z_{t,c}$ to a distribution over the $V$ spectral codes, and the model is trained with
\begin{equation}
\mathcal{L}_{\mathrm{AR}}
=
-\frac{1}{|\mathcal{P}|}
\sum_{(t,c)\in\mathcal{P}}
\log
p_\theta
\left(
y_{t+1,c}
\mid
z_{t,c}
\right),
\qquad
p_\theta(\cdot\mid z_{t,c})
=
\operatorname{softmax}
\left(
\operatorname{MLP}(z_{t,c})
\right),
\end{equation}
where $\mathcal{P}$ denotes the set of valid non-padded channel--time pairs. The prediction head and tokenizer are used only during pre-training and discarded downstream.

\section{Experimental Setup}
\subsection{Pre-training setup}
\href{}{}\label{sec:pretrain}
\paragraph{Corpus.}
For pre-training, we curate seventeen public EEG datasets comprising $23{,}863$ recordings and $3{,}955$ hours of EEG, spanning clinical monitoring, overnight polysomnography, and task-based laboratory recordings, with montages ranging from six to $128$ electrodes. Because NeurDuo identifies channels by electrode coordinates rather than fixed channel indices, recordings retain their native montages without interpolation to a common channel set. Signals are band-pass and notch filtered, resampled to $256$\,Hz, and segmented into $32$\,s sequences with $50\%$ overlap for backbone pre-training. The corpus is split by subject into $22{,}780$ training and $1{,}083$ validation recordings. Dataset-level composition and preprocessing details are provided in Appendix~\ref{app:pretrain}.

\paragraph{Optimisation.} The spectral tokenizer and NeurDuo are both optimized with AdamW and cosine learning-rate decay. NeurDuo is trained from scratch on the full corpus for 16 epochs with an effective batch size of 128 sequences using bfloat16 mixed precision on 32 GPUs. Full optimisation hyperparameters for the tokenizer and NeurDuo are provided in Appendix~\ref{app:pretrain}.

\paragraph{Model family.}
\begin{wraptable}{r}{0.52\columnwidth}
    \centering
    \vspace{-0.8\baselineskip}
    \caption{\textbf{NeurDuo model configurations.}}
    \label{tab:family}
    \fontsize{7}{9}\selectfont
    \renewcommand{\arraystretch}{1.08}
    \setlength{\tabcolsep}{3.5pt}
    \begin{tabular}{lcccccc}
        \toprule
        \vspace{-2pt}
        Model & $H$ & $d_f$ & $L_f$ & $d_s$ & $L_s$ & Params \\
        \midrule
        Small & 128 & 256 & 6  & 192 & 3 & 4.70M \\
        Base  & 192 & 480 & 10 & 320 & 5 & 24.31M \\
        Large & 256 & 672 & 15 & 448 & 7 & 67.83M \\
        \bottomrule
    \end{tabular}
    \vspace{-0.5\baselineskip}
\end{wraptable}

The Small, Base, and Large variants, all pre-trained on 32 A100 GPUs, differ only in model scale: the signal patch encoder width $H$ and the fast- and slow-stream widths and depths $(d_f,L_f)$ and $(d_s,L_s)$ increase together.

\subsection{Downstream Evaluation}
\label{sec:downstream}

\paragraph{Benchmarks and baselines.}
We evaluate NeurDuo on three window tasks, comprising emotion recognition (FACED) \citep{chen2023large}, error-related negativity detection (KaggleERN) \citep{inria-bci-challenge,margaux2012objective}, and vigilance regression (SEED-VIG) \citep{1741-2552-14-2-026017}, and two sequence tasks, comprising sleep staging (Sleep-EDF) \citep{kemp2000analysis,goldberger2000physiobank} and seizure detection (CHB-MIT) \citep{goldberger2000physiobank,shoeb2009application}. None of these datasets appears in our pre-training corpus. We compare against six pre-trained EEG models (BIOT \citep{yang2023biot}, CBraMod \citep{wang2025cbramod}, LaBraM-Base \citep{jiang2024large}, EEGPT-Large \citep{wang2024eegpt}, ST-EEGFormer-S \citep{yang2026eeg}, and REVE-Base \citep{el2026reve}) and three classical pipelines (BP-GBDT \citep{vallat2021open,kastrati2025eeg}, FBCov-TS-Lin \citep{jayaram2018moabb,sabbagh2019manifold}, and ERP-Lin \citep{yang2026eeg}). Dataset and baseline details are provided in Appendices~\ref{app:downstream_task} and~\ref{app:overlap}, respectively. CaMBRAIN~\citep{CaMBRAIN} is omitted because pre-trained weights were unavailable.

\paragraph{Evaluation protocol.}
All foundation models are fine-tuned end-to-end with a common optimisation recipe and a shared readout with matched capacity. By default, sequence samples contain twenty consecutive $30$\,s units ($10$\,min), processed by the same two-layer causal Mamba context head. Baseline backbones encode units independently; NeurDuo carries its fast, slow, and memory-bank states across units. Protocol details and the state-reset comparison are in Table~\ref{tab:state_carrying} and Appendix~\ref{app:downstream}.

\paragraph{Metrics and reporting.}
We report the mean $\pm$ standard deviation over five seeds for FACED and SEED-VIG, four prescribed training folds for KaggleERN, and five subject-disjoint folds for Sleep-EDF and five case-level folds for CHB-MIT, where one subject contributed two case folders (Appendix~\ref{app:chbmit}). Deterministic classical results on FACED and SEED-VIG have no s.d.\ $\mathrm{FP}\cdot\mathrm{h}^{-1}$ counts merged false alarms per recording hour at test-selected thresholds targeting at least $50\%$ unit-level sensitivity (Appendix \ref{app:chbmit}), and is reported without s.d.\ Model size counts backbone parameters. \textbf{Bold} and \underline{underlined} scores mark the best and second-best results per column; arrows indicate metric direction.

\setcounter{topnumber}{1}
\section{Experimental Results}
\subsection{Window tasks}
\textbf{NeurDuo leads all three short-window benchmarks.} We first evaluate NeurDuo on three conventional short-window EEG tasks spanning emotion recognition, error-related negativity detection, and vigilance estimation. NeurDuo achieves the strongest overall performance across all three tasks (Table \ref{tab:window}). On FACED, NeurDuo-Small already outperforms every baseline except the much larger REVE-Base, while NeurDuo-Base further attains the best balanced accuracy and Cohen's $\kappa$. KaggleERN presents a different setting, where the classical ERP-Lin baseline is particularly strong, achieving higher ROC-AUC than every pre-trained baseline; NeurDuo is the only model family to surpass it, with the Small model already achieving the best result. On SEED-VIG, foundation models consistently outperform the classical pipelines, yet NeurDuo again performs best, with NeurDuo-Base improving $R^2$ from $0.297$ for the strongest baseline to $0.364$. These results establish NeurDuo as a strong short-window EEG model across diverse downstream settings, with much of this capability already present in the $4.7$M-parameter Small model.

\begin{table*}[t]
\centering
\caption{\textbf{Short-window downstream performance.} $^\dagger$LaBraM was pre-trained on KaggleERN; see Appendix~\ref{app:overlap}.}
\label{tab:window}
\setlength{\tabcolsep}{1.5pt}
\renewcommand{\arraystretch}{1.08}
\begingroup
\fontsize{7}{9}\selectfont
\medmuskip=1mu
\begin{tabular}{@{}l c c c c c c c@{}}
\toprule
Methods & Model Size & \multicolumn{2}{c}{FACED} & \multicolumn{2}{c}{KaggleERN} & \multicolumn{2}{c}{SEED-VIG} \\
\cmidrule(lr){3-4}\cmidrule(lr){5-6}\cmidrule(lr){7-8}
& & Bal. Acc. $\uparrow$ & Cohen's $\kappa$ $\uparrow$ & ROC-AUC $\uparrow$ & AUC-PR $\uparrow$ & Pearson's $r$ $\uparrow$ & $R^2$ $\uparrow$ \\
\midrule
\textit{BP-GBDT} & -- & 0.173 & 0.070 & $0.557 \pm 0.005$ & $0.753 \pm 0.009$ & 0.445 & 0.031 \\
\textit{FBCov-TS-Lin} & -- & 0.185 & 0.084 & $0.613 \pm 0.023$ & $0.778 \pm 0.023$ & 0.344 & 0.110 \\
\textit{ERP-Lin} & -- & 0.229 & 0.130 & $0.623 \pm 0.019$ & $0.782 \pm 0.012$ & 0.020 & $-0.019$ \\
\midrule
BIOT & 3.2M & $0.364 \pm 0.016$ & $0.283 \pm 0.017$ & $0.496 \pm 0.030$ & $0.708 \pm 0.023$ & $0.535 \pm 0.065$ & $0.179 \pm 0.136$ \\
CBraMod & 4.9M & $0.537 \pm 0.008$ & $0.476 \pm 0.010$ & $0.571 \pm 0.015$ & $0.755 \pm 0.009$ & $0.529 \pm 0.038$ & $0.196 \pm 0.051$ \\
LaBraM $^\dagger$ & 5.8M & $0.555 \pm 0.008$ & $0.496 \pm 0.009$ & $0.546 \pm 0.018$ & $0.737 \pm 0.009$ & $0.588 \pm 0.037$ & $0.188 \pm 0.107$ \\
EEGPT & 25.3M & $0.506 \pm 0.012$ & $0.442 \pm 0.012$ & $0.600 \pm 0.014$ & $0.782 \pm 0.014$ & $0.619 \pm 0.018$ & $0.244 \pm 0.071$ \\
ST-EEGFormer-S & 25.4M & $0.486 \pm 0.007$ & $0.419 \pm 0.008$ & $0.514 \pm 0.047$ & $0.721 \pm 0.022$ & $0.591 \pm 0.032$ & $0.233 \pm 0.067$ \\
REVE-Base & 69.2M & $0.605 \pm 0.007$ & $0.553 \pm 0.008$ & $0.604 \pm 0.018$ & $0.786 \pm 0.024$ & $0.621 \pm 0.020$ & $0.297 \pm 0.023$ \\
\midrule
\rowcolor{black!5}
\textbf{NeurDuo (Small)} & 4.7M & $0.589 \pm 0.005$ & $0.534 \pm 0.006$ & $\mathbf{0.636 \pm 0.014}$ & $\mathbf{0.803 \pm 0.019}$ & $0.632 \pm 0.008$ & $\underline{0.333 \pm 0.005}$ \\
\rowcolor{black!5}
\textbf{NeurDuo (Base)} & 24.3M & $\mathbf{0.610 \pm 0.007}$ & $\mathbf{0.558 \pm 0.007}$ & $0.621 \pm 0.024$ & $0.785 \pm 0.018$ & $\mathbf{0.665 \pm 0.024}$ & $\mathbf{0.364 \pm 0.040}$ \\
\rowcolor{black!5}
\textbf{NeurDuo (Large)} & 67.8M & $\underline{0.607 \pm 0.008}$ & $\underline{0.555 \pm 0.009}$ & $\underline{0.632 \pm 0.010}$ & $\underline{0.790 \pm 0.020}$ & $\underline{0.647 \pm 0.015}$ & $0.294 \pm 0.014$ \\
\bottomrule
\end{tabular}
\endgroup
\end{table*}

\subsection{Sequence tasks}
\label{sec:sequencetask}

\begin{table}[!b]
    \centering
    \caption{\textbf{Long-sequence downstream performance.} CHB-MIT is evaluated using either all $15$ channels or only T7 and P8. $^\dagger$BIOT was pre-trained on SHHS and CHB-MIT; see Appendix~\ref{app:overlap}.}
\label{tab:long_sequence_results}
    \setlength{\tabcolsep}{2.5pt}
    \renewcommand{\arraystretch}{1.08}

    \begingroup
    \fontsize{7}{9}\selectfont
    \begin{tabular}{@{}lccccccc@{}}
        \toprule
        \textbf{Methods} &
        \textbf{Model Size} &
        \multicolumn{2}{c}{\textbf{Sleep-EDF}} &
        \multicolumn{2}{c}{\textbf{CHB-MIT (15 ch)}} &
        \multicolumn{2}{c}{\textbf{CHB-MIT (2 ch)}} \\
        \cmidrule(lr){3-4}\cmidrule(lr){5-6}\cmidrule(lr){7-8}

        & &
        \textbf{Macro-F1} $\uparrow$ &
        \textbf{Cohen's $\kappa$} $\uparrow$ &
        \textbf{AUC-PR} $\uparrow$ &
        \textbf{FP$\cdot$h$^{-1}$} $\downarrow$ &
        \textbf{AUC-PR} $\uparrow$ &
        \textbf{FP$\cdot$h$^{-1}$} $\downarrow$ \\
        \midrule

        \textit{BP-GBDT}
        & -- & $0.710 \pm 0.013$ & $0.695 \pm 0.021$
        & $0.209 \pm 0.109$ & $3.34$ & $0.208 \pm 0.057$ & $3.28$ \\

        \textit{FBCov-TS-Lin}
        & -- & $0.674 \pm 0.018$ & $0.666 \pm 0.019$
        & $0.285 \pm 0.136$ & $2.14$ & $0.224 \pm 0.124$ & $4.12$ \\

        \textit{ERP-Lin}
        & -- & $0.306 \pm 0.044$ & $0.255 \pm 0.031$
        & $0.006 \pm 0.003$ & $15.42$ & $0.009 \pm 0.007$ & $12.65$ \\

        \midrule

        BIOT$^{\dagger}$
        & 3.2M & $0.731 \pm 0.017$ & $0.733 \pm 0.017$
        & $0.208 \pm 0.059$ & $4.73$ & $0.124 \pm 0.094$ & $6.12$ \\

        CBraMod
        & 4.9M & $\underline{0.759 \pm 0.009}$ & $0.749 \pm 0.014$
        & $0.163 \pm 0.085$ & $4.48$ & $0.277 \pm 0.150$ & $3.40$ \\

        LaBraM
        & 5.8M & $\mathbf{0.766 \pm 0.010}$ & $\mathbf{0.758 \pm 0.009}$
        & $0.154 \pm 0.129$ & $12.77$ & $0.056 \pm 0.068$ & $13.41$ \\

        EEGPT
        & 25.3M & $0.739 \pm 0.014$ & $0.729 \pm 0.016$
        & $0.180 \pm 0.079$ & $9.27$ & $0.183 \pm 0.094$ & $7.73$ \\

        ST-EEGFormer-S
        & 25.4M & $0.715 \pm 0.024$ & $0.722 \pm 0.030$
        & $0.034 \pm 0.045$ & $19.71$ & $0.033 \pm 0.025$ & $18.14$ \\

        REVE-Base
        & 69.2M & $0.741 \pm 0.012$ & $0.743 \pm 0.017$
        & $0.163 \pm 0.066$ & $7.31$ & $0.131 \pm 0.104$ & $7.01$ \\

        \midrule
        \rowcolor{black!5}
        \textbf{NeurDuo (Small)}
        & 4.7M & $0.755 \pm 0.008$ & $0.749 \pm 0.011$
        & $\underline{0.413 \pm 0.174}$ & $\mathbf{0.96}$ & $0.370 \pm 0.177$ & $\mathbf{1.62}$
        \\
        \rowcolor{black!5}
        \textbf{NeurDuo (Base)}
        & 24.3M & $0.746 \pm 0.017$ & $0.743 \pm 0.012$
        & $0.392 \pm 0.104$ & $1.34$ & $\underline{0.374 \pm 0.145}$ & $\underline{1.64}$ \\
        \rowcolor{black!5}
        \textbf{NeurDuo (Large)}
        & 67.8M & $0.753 \pm 0.007$ & $\underline{0.752 \pm 0.009}$
        & $\mathbf{0.471 \pm 0.101}$ & $\underline{1.02}$ & $\mathbf{0.429 \pm 0.106}$ & $3.95$ \\
        \bottomrule
    \end{tabular}
    \endgroup
\end{table}

\textbf{NeurDuo excels at seizure detection and remains competitive on sleep staging.} We next turn to the central evaluation setting of this work: long-sequence EEG modelling. We benchmark NeurDuo on Sleep-EDF and CHB-MIT, two sequence-labelling tasks built from ordered EEG units. Table~\ref{tab:long_sequence_results} summarizes the results.
On Sleep-EDF, LaBraM achieves the best performance, while NeurDuo remains competitive and all foundation models fall within a relatively narrow range. One possible reason is that Sleep-EDF provides only two EEG derivations, which leaves little cross-channel structure for NeurDuo's channel-parallel design to exploit.
In contrast, CHB-MIT shows a much clearer separation between our model and the baselines. Using the full 15-channel montage, all NeurDuo variants outperform every baseline on both AUC-PR and false alarms per hour. NeurDuo-Small already reaches an AUC-PR of $0.413$ with $0.96$ false alarms per hour at only $4.7$M parameters, compared with $0.285$ for the strongest non-NeurDuo baseline, while NeurDuo-Large further improves AUC-PR to $0.471$.

\textbf{NeurDuo retains its AUC-PR advantage with only two channels.} To examine performance with fewer channels, we repeat the CHB-MIT evaluation using only T7 and P8, with the same splits and the montage-specific readout budgets in Table \ref{tab:budget}. NeurDuo retains its AUC-PR advantage: the strongest baseline reaches $0.277$, compared with $0.370$ for Small and $0.429$ for Large. Small and Base also give the lowest false-alarm rates in this setting, $1.62$ and $1.64$ per hour. All NeurDuo variants have lower AUC-PR than with 15 channels, consistent with a benefit from richer spatial information, although the readout budget also changes.

\textbf{Longer context helps, with task and montage dependent gains.} 
We evaluate NeurDuo-Small with context lengths from $30$\,s to $60$\,min using the training protocol of Appendix \ref{app:optimisation} (Results in Table \ref{tab:context_ladder}). Sleep-EDF benefits overall from longer context, reaching its highest macro-F1 at $60$\,min. On 15-channel CHB-MIT, AUC-PR peaks at $10$\,min and false alarms are lowest at $30$\,min; both worsen at $60$\,min. With two channels, AUC-PR is highest at $60$\,min, while false alarms are lowest at $30$\,min. The montage moves where that optimum sits: with fifteen channels AUC-PR is already past its best by $30$\,min, whereas with two it is still rising at $60$. Longer causal context is therefore useful, with the preferred length depending on the task, the montage and the metric.

\begin{table}[t]
    \centering
    \caption{\textbf{Effect of causal context length on NeurDuo (Small).} $L$ denotes the number of consecutive $30$\,s units; all other evaluation settings remain fixed. $^\dagger L=1$ provides no cross-unit context.}
\label{tab:context_ladder}
    \setlength{\tabcolsep}{3.5pt}
    \renewcommand{\arraystretch}{1.08}

    \begingroup
    \fontsize{7}{9}\selectfont
    \begin{tabular}{@{}lccccccc@{}}
        \toprule
        \textbf{Context} & $L$ &
        \multicolumn{2}{c}{\textbf{Sleep-EDF}} &
        \multicolumn{2}{c}{\textbf{CHB-MIT (15 ch)}} &
        \multicolumn{2}{c}{\textbf{CHB-MIT (2 ch)}} \\
        \cmidrule(lr){3-4}\cmidrule(lr){5-6}\cmidrule(lr){7-8}
        & &
        \textbf{Macro-F1} $\uparrow$ & \textbf{Cohen's $\kappa$} $\uparrow$ &
        \textbf{AUC-PR} $\uparrow$ & \textbf{FP$\cdot$h$^{-1}$} $\downarrow$ &
        \textbf{AUC-PR} $\uparrow$ & \textbf{FP$\cdot$h$^{-1}$} $\downarrow$ \\
        \midrule

        $30$\,s$^{\dagger}$ & 1
        & $0.696 \pm 0.010$ & $0.695 \pm 0.011$
        & $0.324 \pm 0.095$ & $2.02$ & $0.241 \pm 0.188$ & $7.96$
        \\

        $1$\,min & 2
        & $0.707 \pm 0.009$ & $0.702 \pm 0.010$
        & $0.322 \pm 0.153$ & $4.12$ & $0.290 \pm 0.153$ & $4.05$
        \\

        $2$\,min & 4
        & $0.725 \pm 0.009$ & $0.722 \pm 0.010$
        & $0.404 \pm 0.160$ & $5.83$ & $0.358 \pm 0.137$ & $1.64$
        \\

        $5$\,min & 10
        & $0.744 \pm 0.015$ & $0.742 \pm 0.005$
        & $0.331 \pm 0.128$ & $0.80$ & $0.375 \pm 0.156$ & $2.18$
        \\

        $10$\,min & 20
        & $\underline{0.755 \pm 0.008}$ & $0.749 \pm 0.011$
        & $\mathbf{0.413 \pm 0.174}$ & $0.96$ & $0.370 \pm 0.177$ & $1.62$
        \\

        $30$\,min & 60
        & $0.753 \pm 0.016$ & $\underline{0.752 \pm 0.011}$
        & $\underline{0.406 \pm 0.065}$ & $\mathbf{0.36}$ & $\underline{0.381 \pm 0.173}$ & $\mathbf{1.07}$
        \\

        $60$\,min & 120
        & $\mathbf{0.759 \pm 0.010}$ & $\mathbf{0.755 \pm 0.011}$
        & $0.379 \pm 0.108$ & $\underline{0.56}$ & $\mathbf{0.382 \pm 0.180}$ & $\underline{1.13}$
        \\
        \bottomrule
    \end{tabular}
    \endgroup
\end{table}

\begin{wrapfigure}{r}{0.52\columnwidth}
\vspace{-12pt}
\centering
\includegraphics[width=\linewidth]{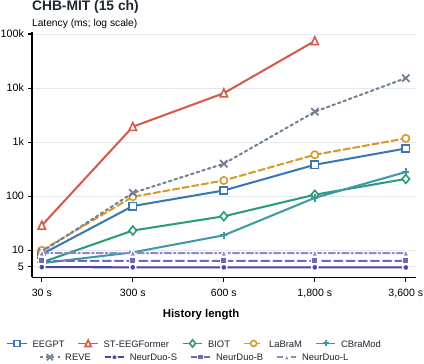}
\caption{\textbf{Inference latency on CHB-MIT.}}
\label{fig:cost}
\vspace{-12pt}
\end{wrapfigure}

\subsection{Streaming Cost}
\label{sec:cost}
\textbf{Per-chunk latency remains nearly constant as history grows.} To verify that NeurDuo can condition on long histories with bounded deployment cost, we measured inference latency as history grows from 30\,s to 1\,h and peak GPU memory at 1,800\,s on two datasets. For each baseline, inference re-encodes the full history window, whereas NeurDuo processes only the next 0.5-s chunk after the history has been streamed. As shown in Figure~\ref{fig:cost}, baseline latency increases with context length, while NeurDuo remains at $4.83$--$8.94$\,ms per chunk and requires only $0.05$--$0.37$\,GB of peak memory. These results demonstrate that NeurDuo supports long-context, real-time EEG processing with nearly constant per-chunk computation and a small memory footprint. Full measurements are provided in Appendix~\ref{app:cost}.

\subsection{Ablation Experiments}
\label{sec:ablation}
We evaluate the contributions of NeurDuo's architectural components on NeurDuo (Small). Table~\ref{tab:ablations} reports architectural ablations and a temporal-order control, with all architectural variants re-pre-trained from scratch under the same settings. Table~\ref{tab:state_carrying} isolates the contribution of persistent backbone state by comparing state carrying across successive units with resets at every 30-s boundary.

\begin{table}[!htb]
\centering
\caption{\textbf{Component ablations and temporal-order control for NeurDuo (Small).} Shuffling is not applicable to FACED, whose samples are independent windows. 
}
\label{tab:ablations}
\begingroup
\fontsize{7}{9}\selectfont
\setlength{\tabcolsep}{2.5pt}
\renewcommand{\arraystretch}{1.08}
\begin{tabular}{@{}l cc cc cc@{}}
\toprule
\textbf{Variant} & \multicolumn{2}{c}{\textbf{FACED}}
& \multicolumn{2}{c}{\textbf{Sleep-EDF}}
& \multicolumn{2}{c}{\textbf{CHB-MIT (15 ch)}} \\
\cmidrule(lr){2-3}\cmidrule(lr){4-5}\cmidrule(lr){6-7}
& Bal.\ Acc.\ $\uparrow$ & $\kappa$ $\uparrow$
& Macro-F1 $\uparrow$ & $\kappa$ $\uparrow$
& AUC-PR $\uparrow$ & FP$\cdot$h$^{-1}$ $\downarrow$ \\
\midrule
\rowcolor{black!5}
\textbf{NeurDuo (Small)}
& $0.589 \pm 0.005$ & $0.534 \pm 0.006$
& $0.755 \pm 0.008$ & $0.749 \pm 0.011$
& $0.413 \pm 0.174$ & $0.96$ \\
\midrule
w/o per-channel tokens
& $0.460 \pm 0.011$ & $0.391 \pm 0.012$ & $0.745 \pm 0.011$ & $0.739 \pm 0.008$
& $0.412 \pm 0.096$ & $1.10$ \\
w/o channel attention
& $0.523 \pm 0.009$ & $0.460 \pm 0.011$ & $0.753 \pm 0.016$ & $0.750 \pm 0.009$
& $0.340 \pm 0.125$ & $4.24$ \\
w/o memory retrieval
& $0.506 \pm 0.012$ & $0.442 \pm 0.013$ & $0.744 \pm 0.012$ & $0.736 \pm 0.010$
& $0.389 \pm 0.112$ & $1.08$ \\
w/o slow stream \& memory
& $0.545 \pm 0.015$ & $0.485 \pm 0.017$ & $0.742 \pm 0.017$ & $0.738 \pm 0.010$
& $0.394 \pm 0.179$ & $0.47$ \\
\midrule
Mamba-only
& $0.523 \pm 0.007$ & $0.461 \pm 0.009$ & $0.744 \pm 0.013$ & $0.737 \pm 0.016$
& $0.406 \pm 0.168$ & $4.60$ \\
\midrule
Shuffled unit order
& --- & --- & $0.735 \pm 0.017$ & $0.733 \pm 0.013$
& $0.371 \pm 0.161$ & $3.79$ \\
\bottomrule
\end{tabular}
\endgroup
\end{table}

\textbf{Channel-specific tokens and channel attention provide complementary benefits.} Removing per-channel tokens degrades performance across all three datasets, with the largest effect on FACED, where Cohen's $\kappa$ drops from $0.534$ to $0.391$. 
The smaller changes on Sleep-EDF and CHB-MIT suggest that per-channel tokens may be more beneficial for short-window tasks such as FACED. 
Channel attention shows a different pattern: removing it has its clearest effect on CHB-MIT, reducing AUC-PR from $0.413$ to $0.340$, while performance on two-channel Sleep-EDF changes little. This is consistent with Section~\ref{sec:sequencetask}, where richer multichannel input provides greater scope for cross-channel interaction.

\textbf{Explicit retrieval is particularly beneficial.}
Removing either memory retrieval or the slow stream with memory lowers Cohen's $\kappa$ on FACED and Sleep-EDF, as well as AUC-PR on CHB-MIT, supporting the role of both components in downstream prediction. This suggests that explicit retrieval strengthens short-window representations by providing access to recent summaries, which may explain its larger effect on FACED.
These benefits do not extend uniformly to every metric: removing the slow stream with memory reduces CHB-MIT false alarms from $0.96$ to $0.47$ per hour, alongside a decrease in AUC-PR from $0.413$ to $0.394$. Mamba-only jointly removes channel attention and the slow stream with memory, retaining only the per-channel fast Mamba backbone. It performs worse across all reported metrics, showing that the full multi-component architecture consistently outperforms the per-channel Mamba backbone alone.
Shuffling unit order also degrades Sleep-EDF and CHB-MIT performance, indicating that temporal order contributes to sequence modelling.

\begin{wraptable}{r}{0.56\textwidth}
\vspace{-\intextsep}
\centering
\caption{\textbf{Persistent state in NeurDuo (Small).} Backbone state is reset every 30 s or carried across a 10 min sequence, keeping the context-head architecture fixed.}
\label{tab:state_carrying}
\fontsize{7}{9}\selectfont
\setlength{\tabcolsep}{2pt}
\renewcommand{\arraystretch}{1.08}
\begin{tabular}{@{}lcccc@{}}
\toprule
\textbf{State} & \multicolumn{2}{c}{\textbf{Sleep-EDF}}
& \multicolumn{2}{c}{\textbf{CHB-MIT (15 ch)}} \\
\cmidrule(lr){2-3}\cmidrule(lr){4-5}
& Macro-F1 $\uparrow$ & $\kappa$ $\uparrow$
& AUC-PR $\uparrow$ & FP$\cdot$h$^{-1}$ $\downarrow$ \\
\midrule
Reset & $0.745 \pm 0.014$ & $0.738 \pm 0.013$
& $0.307 \pm 0.191$ & $5.16$ \\
\rowcolor{black!5}
Carry & $\mathbf{0.755 \pm 0.008}$ & $\mathbf{0.749 \pm 0.011}$
& $\mathbf{0.413 \pm 0.174}$ & $\mathbf{0.96}$ \\
\bottomrule
\end{tabular}
\end{wraptable}

\textbf{Persistent backbone state improves both sequence tasks for NeurDuo-Small.} Table~\ref{tab:state_carrying} compares carrying the backbone state across the full 10-min sequence with resetting it at every 30-s boundary, while keeping the downstream sequence head unchanged. Carrying state improves performance on both tasks, with a modest gain on Sleep-EDF and a larger improvement on CHB-MIT, where AUC-PR increases from $0.307$ to $0.413$. These results show that information retained within the NeurDuo backbone across successive units contributes directly to downstream prediction.

\textbf{Pre-training improves performance on all five tasks.} Appendix \ref{app:no_pretrain} shows that, relative to random initialisation, it raises NeurDuo-Small’s FACED balanced accuracy from $0.488$ to $0.589$ and 15-channel CHB-MIT AUC-PR from $0.309$ to $0.413$ under the same fine-tuning protocol. With architecture and readout held fixed, these gains support the value of pre-trained representations across task types. The CHB-MIT improvement further shows that pre-training on $32$\,s sequences remains beneficial when adapting to $10$\,min downstream context.

\FloatBarrier
\section{Conclusion}

This paper introduces NeurDuo, a causal EEG foundation model for continuous multichannel EEG with persistent state and structured explicit memory. Pre-trained on 3,955 hours of EEG, NeurDuo achieves the best performance on four of five downstream benchmarks, including all three short-window tasks and seizure detection, while supporting efficient streaming inference over long histories. These results show that persistent, multi-timescale modelling extends EEG foundation models beyond isolated windows while also surpassing existing baselines on short-window tasks.

Several limitations remain. NeurDuo is trained with a local prediction objective, relies on implicit recurrent state for long-range retention beyond its short explicit memory, and is evaluated on only two continuous long-sequence tasks. Future work will explore longer-horizon objectives, more adaptive memory mechanisms, and broader continuous EEG evaluation. 
Nevertheless, we believe our work points toward a broader shift from isolated-window representation learning to continuous long-horizon modelling through persistent memory management, opening a new direction for foundation models of continuous neural signals.



\bibliography{iclr2027_conference}
\bibliographystyle{iclr2027_conference}

\newpage
\setcounter{topnumber}{2}
\appendix
\section{Model and Pre-training Details}
\label{app:modelandpretraining}
\subsection{Pre-training Details}
\label{app:pretrain}
\paragraph{Corpus.} Tables ~\ref{tab:corpus-domain} and \ref{tab:corpus} give the corpus by domain and by dataset. The composition is uneven by necessity rather than by choice: continuous overnight and clinical recordings are the only sources that exist at scale, and they supply more than nine tenths of the hours, while the montage range is set by the task-based corpora at the other end. Because the metadata encoder is given electrode coordinates and never a dataset label (Appendix~\ref{app:metadata}), the model is given no explicit corpus-identity cue with which to exploit this imbalance.

\begin{table}[!htbp]
\centering
\caption{\textbf{Pre-training corpus by domain.} All counts are after preprocessing. \#Rec.\ is the number of recordings and Hours their total duration; \#Ch.\ is the electrodes retained once channel labels are resolved to standard positions, given as a range where a domain's corpora use different montages; \#Windows is the number of $32$\,s pre-training windows the recordings are cut into, at $50\%$ overlap. Table~\ref{tab:corpus} gives the per-dataset breakdown.}
\label{tab:corpus-domain}
\begingroup
\fontsize{7}{9}\selectfont
\setlength{\tabcolsep}{2.5pt}
\renewcommand{\arraystretch}{1.08}
\begin{tabular}{@{}llrrcr@{}}
\toprule
\textbf{Domain} & \textbf{Datasets}
& \textbf{\#Rec.} & \textbf{Hours}
& \textbf{\#Ch.} & \textbf{\#Windows} \\
\midrule
Clinical
& TUSZ, TUEP, Siena, TUAR
& 8{,}007 & 1{,}811.9 & 17--19 & 395{,}424 \\
Sleep (PSG)
& PhysioNet-2018, ISRUC
& 11{,}339 & 1{,}882.3 & 6 & 406{,}517 \\
Motor
& EEGMMIDB, HGD, GAL, BCI IV-1
& 1{,}821 & 95.5 & 32--126 & 18{,}415 \\
Attention
& Cao 2019
& 516 & 81.9 & 30 & 17{,}652 \\
Emotion
& DEAP, DREAMER
& 1{,}694 & 46.2 & 14--32 & 7{,}367 \\
Speech (auditory)
& Broderick
& 380 & 20.6 & 128 & 4{,}056 \\
ERP
& Brain Invaders
& 64 & 12.4 & 16 & 2{,}698 \\
Resting state
& Trujillo, SPIS
& 42 & 3.8 & 64 & 805 \\
\midrule
\textbf{Total}
& 17 datasets
& \textbf{23{,}863} & \textbf{3{,}954.7}
& \textbf{6--128} & \textbf{852{,}934} \\
\bottomrule
\end{tabular}
\endgroup
\end{table}

\begin{table}[t]
\centering
\caption{\textbf{Pre-training corpus, per dataset, after preprocessing.} The per-dataset expansion of Table~\ref{tab:corpus-domain}, ordered by hours. \#Rec.\ is the number of recordings and Hours their total duration; \#Ch.\ is the electrodes retained once channel labels are resolved to standard positions; \#Windows is the number of $32$\,s pre-training windows the recordings are cut into, at $50\%$ overlap. Where a channel count differs from the one quoted in a release's own publication, the table reports the signal channels the released files contain.}
\label{tab:corpus}
\begingroup
\fontsize{7}{9}\selectfont
\setlength{\tabcolsep}{2.5pt}
\renewcommand{\arraystretch}{1.08}
\begin{tabular}{@{}llrrcr@{}}
\toprule
\textbf{Dataset} & \textbf{Domain}
& \textbf{\#Rec.} & \textbf{Hours}
& \textbf{\#Ch.} & \textbf{\#Windows} \\
\midrule
PhysioNet-2018
\tabcite{ghassemi2018you,goldberger2000physiobank}
& Sleep (PSG) & 6{,}830 & 1{,}138.3 & 6 & 245{,}880 \\

TUSZ \tabcite{shah2018temple}
& Clinical & 4{,}959 & 911.2 & 17--19 & 197{,}388 \\

ISRUC \tabcite{khalighi2016isruc}
& Sleep (PSG) & 4{,}509 & 744.0 & 6 & 160{,}637 \\

TUEP \tabcite{veloso2017big}
& Clinical & 2{,}697 & 631.7 & 17--19 & 138{,}035 \\

Siena \tabcite{detti2020eeg}
& Clinical & 41 & 169.0 & 19 & 37{,}963 \\

TUAR \tabcite{hamid2020temple}
& Clinical & 310 & 100.0 & 19 & 22{,}038 \\

Cao 2019 \tabcite{cao2019multi}
& Attention & 516 & 81.9 & 30 & 17{,}652 \\

EEGMMIDB
\tabcite{schalk2004bci2000,goldberger2000physiobank}
& Motor & 1{,}526 & 48.5 & 64 & 8{,}282 \\

HGD \tabcite{schirrmeister2017deep}
& Motor & 185 & 28.6 & 126 & 6{,}161 \\

DREAMER \tabcite{katsigiannis2017dreamer}
& Emotion & 414 & 23.8 & 14 & 4{,}807 \\

DEAP \tabcite{koelstra2011deap}
& Emotion & 1{,}280 & 22.4 & 32 & 2{,}560 \\

Broderick \tabcite{broderick2018electrophysiological}
& Speech (auditory) & 380 & 20.6 & 128 & 4{,}056 \\

Brain Invaders \tabcite{congedo2011brain}
& ERP & 64 & 12.4 & 16 & 2{,}698 \\

GAL \tabcite{luciw2014multi}
& Motor & 96 & 10.0 & 32 & 2{,}102 \\

BCI IV-1
\tabcite{tangermann2012review,blankertz2007non}
& Motor & 14 & 8.4 & 49 & 1{,}870 \\

Trujillo \tabcite{trujillo2017effect}
& Resting state & 22 & 3.0 & 64 & 645 \\

SPIS \tabcite{torkamani2020prediction}
& Resting state & 20 & 0.8 & 64 & 160 \\
\midrule
\textbf{Total}
& 17 datasets
& \textbf{23{,}863}
& \textbf{3{,}954.7}
& \textbf{6--128}
& \textbf{852{,}934} \\
\bottomrule
\end{tabular}
\endgroup
\end{table}

\paragraph{Preprocessing and windowing.} Each recording is band-pass filtered to $0.5$--$75$\,Hz, notch filtered, and resampled to $256$\,Hz. Channel labels are resolved against a standard electrode dictionary and mapped to coordinates on the unit sphere; channels that cannot be resolved are dropped, which is why the polysomnographic sources retain only their EEG derivations. Recordings are then cut into sequences of $64$ chunks of $128$ samples: $32$\,s, with chunks non-overlapping, taken at a stride of $32$ chunks for backbone pre-training, so consecutive sequences overlap by half; the tokenizer of stage 1 is trained on the same sequences at a stride of $64$, without overlap. Each sequence is an independent training sample: the recurrent state is initialised at its start and discarded at its end, so no gradient crosses a sequence boundary during pre-training. The half-sequence stride is not merely a way of drawing more samples from the same hours. Because state is reset at every boundary, a given moment of a recording would otherwise always fall at the same offset within a sequence, and so would always be seen with the same amount of history behind it; at a stride of half a sequence it falls at two offsets $32$ chunks apart, and the model is required to predict from a nearly cold state and from a warm one on the same signal. A transition shorter than the stride is also left uncut in at least one sequence. Downstream sequences are not overlapped: there the window is fixed by the labelling unit, a scoring epoch, an annotation interval, an event-locked trial, and the warm-up is instead absorbed by state passing from one unit to the next within a sequence.

\begin{table}[!t]
\centering
\caption{\textbf{Pre-training hyperparameters.} Stage 1 trains the spectral tokenizer and stage 2 the backbone, with the tokenizer frozen between them. A value spanning both columns is shared by the two stages; the Tokenizer block describes the stage-1 model itself and so is not stage-dependent. Every entry is identical across the three model sizes except the peak learning rate, given as S/B/L. ``--'' marks a setting that does not apply to that stage.}
\label{tab:hparams}
\begingroup
\fontsize{7}{9}\selectfont
\setlength{\tabcolsep}{2.5pt}
\renewcommand{\arraystretch}{1.08}
\begin{tabular}{@{}llll@{}}
\toprule
& \textbf{Hyperparameter}
& \textbf{Stage 1 (tokenizer)}
& \textbf{Stage 2 (backbone)} \\
\midrule
\multirow{3}{*}{Data}
& Chunk
& \multicolumn{2}{l}{$128$ samples ($0.5$\,s), non-overlapping} \\
& Sequence length
& $64$ chunks ($32$\,s)
& $64$ chunks ($32$\,s) \\
& Sequence stride
& $64$ (no overlap)
& $32$ ($50\%$ overlap) \\
\midrule
\multirow{14}{*}{Optimisation}
& Optimiser
& AdamW
& AdamW \\
& Peak learning rate
& $5\times10^{-5}$
& $3.0/2.2/1.9\times10^{-4}$ (S/B/L) \\
& Weight decay
& $10^{-4}$
& $0.05$ \\
& Schedule
& cosine to $0.05\times$
& cosine to $0.05\times$ \\
& Warm-up
& $500$ steps
& $1{,}000$ steps \\
& Gradient clipping
& $1.0$
& $1.0$ \\
& Frontend LR multiplier
& ---
& $0.1$ \\
& Adam $(\beta_1,\beta_2)$
& $(0.9, 0.999)$
& $(0.9, 0.999)$ \\
& Adam $\epsilon$
& $10^{-8}$
& $10^{-8}$ \\
& Epochs
& $2$
& $16$ \\
& Effective batch
& $64$ sequences
& $128$ sequences \\
& Updates
& $12{,}832$
& $102{,}096$ \\
& Precision
& float16 (loss scaling)
& bfloat16 \\
& Seed
& $1337$
& $1337$ \\
\midrule
\multirow{6}{*}{Tokenizer}
& Input feature
& \multicolumn{2}{l}{$2\times65$ (rFFT amplitude $\oplus$ phase of a $128$-sample chunk)} \\
& Codebook
& \multicolumn{2}{l}{$2{,}048\times64$, $\ell_2$-normalised} \\
& Encoder
& \multicolumn{2}{l}{two pre-norm residual MLP blocks, width $128$} \\
& Decoder width
& \multicolumn{2}{l}{$128$} \\
& Codebook update
& \multicolumn{2}{l}{EMA $0.99$, initialised from first-batch samples, dead-code reset} \\
& Loss weights
& \multicolumn{2}{l}{amplitude $1.0$, phase $1.0$, commitment $\beta=1.0$} \\
\bottomrule
\end{tabular}
\endgroup
\end{table}

\begin{table}[!b]
\centering
\caption{\textbf{Architecture constants.} Identical in all three model sizes; the widths and depths that scale are given in Table~\ref{tab:family}. One step is one $0.5$\,s chunk, so the durations in parentheses follow from the step counts beside them.}
\label{tab:arch}
\begingroup
\fontsize{7}{9}\selectfont
\setlength{\tabcolsep}{2.5pt}
\renewcommand{\arraystretch}{1.08}
\begin{tabular}{@{}lll@{}}
\toprule
& \textbf{Hyperparameter} & \textbf{Value} \\
\midrule
\multirow{3}{*}{Frontend}
& Chunk encoder kernels
& $\{3,7,15\}$, three parallel branches \\
& Metadata encoder width
& $64$ \\
& Channel attention
& every $2$ fast layers, $4$ heads \\
\midrule
\multirow{3}{*}{Streams}
& SSM state dimension
& $16$ (fast and slow) \\
& SSM expansion factor
& $2$ (fast and slow) \\
& Causal convolution kernel
& $4$ (fast), $3$ (slow) \\
\midrule
\multirow{3}{*}{Memory}
& Writer pooling window $w$
& $8$ steps ($4$\,s) \\
& Writer queue slots
& $16$ (ring buffer; each write pools the most recent $w$) \\
& Write interval $N$
& $4$ steps ($2$\,s) \\
& Bank slots $B$
& $8$ ($16$\,s of explicit memory) \\
\midrule
\multirow{3}{*}{Objective}
& Prediction horizon
& $1$ step ($0.5$\,s) \\
& Target vocabulary
& $2{,}048$ codes \\
& Predictor width
& $d_f$ \\
\bottomrule
\end{tabular}
\endgroup
\end{table}

\paragraph{Montages.} No montage unification is applied. Each recording enters training with the channel set its release provides, resolved to coordinates, but neither reduced to a common subset nor interpolated onto one. The seventeen corpora contribute recordings of $6$, $14$, $16$, $17$, $19$, $30$, $32$, $49$, $64$, $126$ and $128$ channels, and recordings of different widths are drawn into the same batch. Batching pads the channel axis to the widest recording in the batch and records the padding in a mask; padded channels are excluded from the loss and masked out of channel attention, so they influence no other channel. The $6$--$128$ range in Table~\ref{tab:corpus} is therefore a description of the corpus rather than a supported operating range: because a channel is addressed by its coordinate and not by its position in a fixed montage (\S\ref{sec:coordinateAware}), how many channels a recording has is not a quantity the architecture has to be told. Where the count in the table differs from the one quoted in a release's original publication, the table reports the signal channels the released files contain; \cite{cao2019multi}, for example, is recorded with a $32$-electrode cap of which two are mastoid references, and ships $30$ EEG channels.

\paragraph{Hyperparameters.} Table~\ref{tab:hparams} lists both stages and Table~\ref{tab:arch} the architecture constants that Table~\ref{tab:family} does not carry. Two implementation choices are particularly relevant to training stability. The SSM parameters $A_{\log}$ and $D$ are excluded from weight decay along with biases and normalisation scales: $A_{\log}$ parameterises the state decay rates, so decaying it compresses the range of timescales the recurrence can represent. And precision must be bfloat16 rather than float16, which overflows inside the selective scan: a float16 run of an earlier configuration diverged irrecoverably after roughly three thousand updates, and gradient clipping did not prevent it.

\begin{figure}[!t]
\centering
\includegraphics[width=\linewidth]{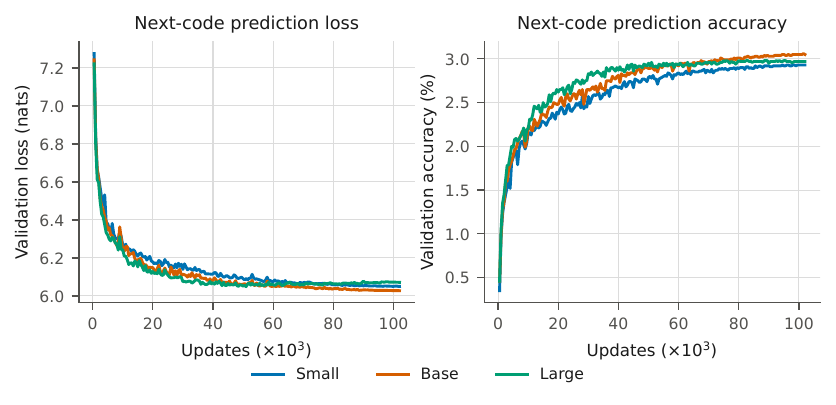}
\caption{\textbf{Pre-training curves for the three sizes of Table~\ref{tab:family}}, on the held-out recordings. Small and Base reach their minima as the schedule ends, at $6.0503$ and $6.0272$ nats; Large reaches its minimum of $6.0491$ after $51{,}000$ updates and then rises to $6.0715$.}
\label{fig:pretrain-curves}
\end{figure}

\paragraph{Training curves.} Figure~\ref{fig:pretrain-curves} plots validation loss and next-code accuracy against updates for the three sizes, which differ only in width, depth and peak learning rate; corpus, schedule, effective batch and number of updates are identical. The loss is a $2{,}048$-way cross-entropy taken per channel and per step, so a uniform predictor sits at $\ln 2{,}048 = 7.62$ nats and the figure's vertical range covers the last $1.6$ nats of that. Accuracy is the top-1 rate on that same $2{,}048$-way choice, for which a uniform predictor scores $0.049\%$; the three runs end at $2.93$, $3.05$ and $2.97\%$. Predicting the exact code half a second ahead remains challenging. The observed improvement reflects temporal structure that makes some future codes more likely than others. The three curves separate early, and their minima then lie within 0.03 nats of one another, which is the observation the model family rests on. Small and Base are still descending at the end of the schedule, but at under $10^{-3}$ nats per thousand updates, so little is left on the table.

\subsection{Metadata: which fields are encoded, and why}
\label{app:metadata}
The metadata encoder of \S\ref{sec:coordinateAware} builds its per-channel bias $e_c$ from two fields. It could in principle use more, so we audited every field our preprocessing pipeline carries, over the training split of the pre-training corpus (\S\ref{sec:pretrain}), $22{,}780$ recordings; Table~\ref{tab:metafields} lists the dispositions. Two fields turn out to be constant and a third nearly so, and would only add the same offset to every sample. One is informative but cannot be used, for the reason given below.

Because the channel-type field is constant, electrode coordinates are the only signal by which the model can tell one channel from another; without them the frontend carries no electrode identity: it is applied to each channel independently, so permuting the channels would simply permute its outputs, and the channel attention of \S\ref{sec:backbone} would receive tokens that differ by their signal alone, with nothing to say which electrode each came from. Coordinates are also what admits a varying montage: a lookup keyed by channel index cannot be shared across corpora whose channel counts range from $6$ to $128$, whereas a position on the scalp means the same thing everywhere.

The reference scheme is kept on different grounds: it describes how a recording was made rather than which recording it is, and unlike corpus identity it is drawn from a small shared vocabulary, so a new dataset maps onto a category the model has already seen.

Corpus identity is excluded because it is not available where it would be used. The pre-training corpus comprises $17$ datasets, so a corpus embedding has $17$ trained rows, while every downstream dataset in this paper is by construction absent from that corpus, which is what makes the evaluation a transfer evaluation. Its embedding row would therefore still be at its random initialisation, and conditioning on it would inject an untrained vector into every token of every downstream sample.

Within pre-training, corpus identity is genuinely informative, and that is the second objection to it: the seventeen corpora differ systematically in montage, recording device and subject population, so a model told which corpus a window came from can lower its pre-training loss by separating corpora rather than by learning the structure they share. A backbone-free probe reduces cross-entropy by about $0.15$ nats relative to a marginal predictor using recording-level code statistics, while extending history beyond $4$\,s adds at most $0.017$ nats. Overlap between windows prevents attributing the former solely to recording identity (Appendix~\ref{app:horizon}). Separately, on CHB-MIT, an oracle told only whether a unit's recording contains a seizure, and never given the signal, reaches AUROC $0.905$, where the best published result, obtained under an easier split, is $0.921$ (Appendix~\ref{app:chbmit}). These observations concern recording-level information rather than corpus identity. We did not train a variant that receives corpus identity, so they motivate the decision rather than constitute an ablation of it.

\begin{table}[!t]
\centering
\caption{\textbf{Metadata fields available in the pipeline and their disposition.}}
\label{tab:metafields}
\begingroup
\fontsize{7}{9}\selectfont
\setlength{\tabcolsep}{2.5pt}
\renewcommand{\arraystretch}{1.08}
\begin{tabular}{@{}llp{6.4cm}@{}}
\toprule
\textbf{Field} & \textbf{Used} & \textbf{Reason} \\
\midrule
Electrode coordinates
& \checkmark
& Physical position; the only cue distinguishing channels. \\
Reference scheme
& \checkmark
& Eight values across the pre-training corpus, reflecting how each recording was referenced. \\
Channel type
& ---
& Constant: all $438{,}323$ channels share one value. \\
Sampling rate
& ---
& Constant: every recording is resampled to $256$\,Hz. \\
Device identifier
& ---
& Near-constant: $22{,}740$ of $22{,}780$ recordings share one value. \\
Corpus identifier
& ---
& Informative but unavailable at transfer time. \\
Missing-channel mask
& ---
& Not encoded as a token, but retained to mask padded channels in the channel attention and exclude them from the codebook statistics. \\
\bottomrule
\end{tabular}
\endgroup
\end{table}

\section{Pre-training Objective Analysis}
\subsection{Why the prediction target is discrete}
\label{app:collapse}

The obvious objective for a causal model is to predict its own representation of the next step. We began there and could not make it work; because the failure survived the remedies we tried rather than yielding to tuning, and because two standard remedies do not help, we record the diagnosis here.

\paragraph{Setup.} These runs predate the corpus and the model family of the main text: an earlier version of the data ($7{,}607$ training and $400$ validation recordings, against the $22{,}780$ and $1{,}083$ of \S\ref{sec:pretrain}) and a much smaller backbone. The unmodified objective ran to completion, $17{,}703$ updates over $20.7$ hours; the three variants are four-hour probes stopped by their wall-clock limit rather than by convergence, so what follows compares trajectories, not converged values. Their configurations differ by a single field each: centring is the only change from the unmodified objective, the variance and covariance weights the only change from the centred run, and the discrete target the only change from it in turn.

\paragraph{The objective and its collapse.} A student model produces a state from the signal seen so far, and a predictor maps that state to the frontend token $k$ steps ahead. The target token comes not from the student itself but from an exponential-moving-average copy of its frontend alone, patch encoder, metadata encoder and channel attention, which receives no gradient. Training is contrastive with in-batch negatives, same-recording negatives excluded. After twenty hours the validation loss had not moved. A pass over the held-out recordings, measuring the cosine similarity between target representations within and across recordings, the fraction of their norm carried by their mean, and the predictor's alignment with the true future against a randomly drawn one, showed why. The cosine similarity between target representations was $1.0000$ both within and across recordings, with $99.99\%$ of their norm carried by a component shared across all inputs. The loss sat at $8.2535 = \log 3841$, the value a contrastive objective takes when its candidates are indistinguishable. The predictor aligned with the true future at $0.1293$ and with a random future at $0.1290$. The student's own frontend output collapsed in the same way, so this is not an artefact of the moving-average copy. It is not a failure of the data either: subtracting the shared component drops the across-recording cosine from $1.0000$ to $0.013$, close to orthogonal, the information was present throughout, dominated by a constant the objective had no reason to remove.

\paragraph{Collapse in our continuous-target configuration.} In our setup the keys come from a momentum-updated copy of the frontend, which receives no gradient of its own and only follows the student. Once the target representations become indistinguishable the contrastive gradient with respect to the query vanishes, which is why the unmodified run sat at $\log 3841$ for twenty hours. Objectives of this shape are trained successfully elsewhere with momentum encoders, so we do not claim collapse must follow; in our configuration training did not recover from it.

\paragraph{Two standard remedies do not fix it.} Centring the representations before normalisation, in the manner of self-distillation methods, repaired the training side convincingly: training loss fell from $8.25$ to between $4.8$ and $5.6$ and training accuracy rose from chance to twenty or thirty percent, while validation loss rose from $8.35$ to $11.0$ and validation accuracy stayed flat. Centring removes the shared component when the loss is evaluated; the representation still consists of that component plus a residual four orders of magnitude smaller. Variance and covariance regularisation made matters worse: the regulariser took effect, its variance term falling from $0.95$ to $0.06$, but validation accuracy halved and validation loss diverged past the level reached without it, and probing on validation data found the target still collapsed at an across-recording cosine of $0.9996$.

\paragraph{A frozen discrete target removed it.} Replacing the continuous target with the frozen tokenizer's codes and the contrastive loss with cross-entropy over the codebook, the only change from the centred run, reverses all three symptoms: validation loss decreases monotonically from $7.609$ to $6.729$ without diverging, validation accuracy rises monotonically instead of remaining flat, and the training--validation gap closes to within a tenth of a percentage point. The mechanism is that the tokenizer reconstructs a fixed external quantity, the spectrum, so the target cannot drift toward a constant of its own; a constant prediction remains available to the backbone, as Appendix~\ref{app:headact} shows for the readout, but it is no longer a configuration the objective cannot move away from, as it was when the target moved with the model. The general lesson is that collapse of this kind lives in the representation, where a regulariser applied at the loss can hide it without removing it; the informative measurement is on the representation itself, not on the loss.

\subsection{How much history a prediction horizon requires}
\label{app:horizon}
The temporal horizon of the pre-training objective (\S\ref{sec:multichannel}) is one step. Because a horizon is only worth extending if the signal carries structure at that range, we measured how far back a predictor must look as a function of how far ahead it must predict. The measurement deliberately does not involve the backbone: a model that never learned to use long history produces a flat curve whether or not the history is informative, so a backbone-based probe cannot separate the signal has no long-range structure from our model failed to exploit it.

\paragraph{Probe.} We freeze the stage-1 tokenizer and convert recordings to code sequences at one code per $0.5$\,s. The $2048$ codebook vectors are clustered into $M=32$ coarse classes by $k$-means, so each step becomes a point in a $32$-way alphabet and each interval becomes a normalised histogram. Given the histogram $X_K$ of the preceding $K$ steps, a linear $M\to M$ model predicts the histogram of a short interval $H$ steps later; the score is held-out soft-label cross-entropy (CE) in nats, lower being better, in the same unit as the pre-training loss. With $M=32$ the predictor has about a thousand parameters and is a deliberately low-capacity reader of $19{,}200$ windows. The probe is split by recording, so no recording contributes to both fitting and evaluation.

\paragraph{Controlling for recording identity.} Longer history can improve this score even when the signal is stationary, because it estimates \emph{that recording's} own code distribution more precisely, and recording identity is exactly the kind of cue that does not transfer across subjects. We therefore do not deform the target distribution (an earlier version subtracted each recording's global histogram from both sides, which pinned every cell at $\ln 32=3.466$ and leaked the target, since that histogram was computed from the window being predicted). Instead we compare two predictors: one given only a leave-one-out estimate $R_{\mathrm{loo}}$ of the recording's code distribution, computed from that recording's \emph{other} windows, and one given $[R_{\mathrm{loo}}, X_K]$. The reported quantity is the difference, $\mathrm{CE}(\text{identity}) - \mathrm{CE}(\text{identity}+K)$: what $K$ seconds of history are worth once that estimate has been accounted for. One limitation of this control is worth stating. Windows are cut at half-window stride, so a neighbouring window may overlap the interval the current one is asked to predict, and $R_{\mathrm{loo}}$ may therefore contain information about the target and not only about the recording. Both predictors receive it, and we do not claim a direction for the residual bias, so we read these numbers as exploratory rather than as a leakage-free estimate of what history contributes. Within each row, all values of $K$ use the same estimate, which holds the conditioning statistic fixed but does not eliminate possible overlap-induced bias in comparisons across history lengths.

\paragraph{Implementation.} The codebook is clustered by $40$ iterations of spherical $k$-means, cosine similarity coinciding with the inner product because the codebook is $\ell_2$-normalised. Each window supplies $256$ past steps; the feature $X_K$ is the normalised histogram of the last $K\in\{8,16,32,64,128,256\}$ of them ($4$ to $128$\,s), and the target $Y_H$ the normalised histogram of the $8$ steps ($4$\,s) beginning $H\in\{2,20,60,120\}$ steps ($1$, $10$, $30$, $60$\,s) after the end of the past region. Each predictor is a single linear layer with zero-initialised weights, trained full-batch for $300$ Adam steps (learning rate $0.05$, weight decay $10^{-4}$) on soft-label cross-entropy, with one fifth of recordings held out. We report the best held-out value reached during that optimisation. These are validation-selected scores, descriptive rather than unbiased estimates, and we read the difference between two predictors selected the same way rather than either score on its own. Of the $19{,}200$ windows collected, $17{,}026$ come from recordings contributing at least two windows and are usable for the leave-one-out estimate.

\begin{table}[!b]
\centering
\caption{\textbf{Exploratory history-probe results conditioned on recording-level statistics.} Rows are prediction horizons and columns the length of history supplied. \textsc{Marginal} is a predictor given no input and \textsc{Identity} one given only a leave-one-out estimate of the recording's own code distribution; the first two columns are their absolute held-out cross-entropies in nats, and the remaining six are $\mathrm{CE}(\textsc{Identity}) - \mathrm{CE}(\textsc{Identity}{+}K)$, the reduction in cross-entropy from adding $K$ seconds of history to the same recording-level estimate. Best denotes the history length $K$ with the largest improvement, also marked in \textbf{bold}. Measured on $19{,}200$ windows drawn from $4{,}600$ pre-training recordings, of which $17{,}026$ windows support the leave-one-out estimate, split by recording.}
\label{tab:horizon}
\begingroup
\fontsize{7}{9}\selectfont
\setlength{\tabcolsep}{2.5pt}
\renewcommand{\arraystretch}{1.08}
\begin{tabular}{@{}lcc cccccc c@{}}
\toprule
& \multicolumn{2}{c}{\textbf{Absolute CE} ($\downarrow$)}
& \multicolumn{6}{c}{\textbf{Improvement over \textsc{identity}} ($\uparrow$) \textbf{for} $K=$}
& \\
\cmidrule(lr){2-3}\cmidrule(lr){4-9}
\textbf{Predict ahead}
& \textbf{\textsc{marginal}}
& \textbf{\textsc{identity}}
& $\mathbf{4}$\,\textbf{s}
& $\mathbf{8}$\,\textbf{s}
& $\mathbf{16}$\,\textbf{s}
& $\mathbf{32}$\,\textbf{s}
& $\mathbf{64}$\,\textbf{s}
& $\mathbf{128}$\,\textbf{s}
& \textbf{Best} \\
\midrule
$1$\,s
& 3.4435 & 3.2911
& 0.1048 & 0.1090 & \textbf{0.1097}
& 0.1070 & 0.1038 & 0.0982
& $16$\,s \\
$10$\,s
& 3.4418 & 3.2918
& 0.0892 & 0.0952 & 0.0988
& \textbf{0.1001} & 0.0984 & 0.0948
& $32$\,s \\
$30$\,s
& 3.4433 & 3.2922
& 0.0761 & 0.0835 & 0.0891
& 0.0915 & \textbf{0.0920} & 0.0895
& $64$\,s \\
$60$\,s
& 3.4398 & 3.2913
& 0.0695 & 0.0756 & 0.0817
& 0.0854 & \textbf{0.0865} & 0.0854
& $64$\,s \\
\bottomrule
\end{tabular}
\endgroup
\end{table}

\paragraph{Predictors compared.} Three predictors are compared for each horizon on the same held-out recordings. \textsc{marginal} emits the mean training target without fitting, and is the highest value in every row of the table. \textsc{identity} and \textsc{identity+K} use the linear-softmax form and the training recipe above, and differ only in what they are given. \textsc{Identity} receives $R_{\mathrm{loo}}$ alone, whereas
\textsc{Identity}$+K$ additionally receives the history histogram $X_K$. Table~\ref{tab:horizon} reports the first two as absolute cross-entropies and the third as its improvement over \textsc{identity}, from which every absolute value can be recovered: for the $1$\,s probe horizon and $16$\,s of history, $3.2911-0.1097=3.1814$.

\paragraph{Scope of the data.} A window must span the $128$\,s past region, the furthest prediction offset and the target interval, so $192$\,s in total, which admits $18{,}866$ of the $22{,}780$ training recordings (\S\ref{sec:pretrain}); the three corpora excluded outright by this filter are trial-structured and too short to carry long-range structure in any case. The surviving pool is dominated by continuous sleep and clinical recordings, four corpora accounting for ninety-two percent of it, so the table below should be read as a statement about long continuous EEG rather than about all EEG, which is also the regime that the long-context claims of \S\ref{sec:sequencetask} concern. Windows are drawn at random from that pool rather than from the head of the manifest.

\paragraph{Magnitudes.} Relative to \textsc{Marginal}, \textsc{Identity} reduces cross-entropy by $0.148$--$0.152$ nats across the four horizons. The best history-conditioned predictor at each horizon achieves a further reduction of $0.086$--$0.110$ nats, while extending history beyond $4$\,s adds $0.005$--$0.017$ nats. These values describe the performance of linear probes on coarse code histograms under the stated conditioning and validation-selection procedure. Given the possible target overlap, they should not be interpreted as an isolated recording-identity effect or a leakage-free estimate of the contribution of history.

\paragraph{Reading.} In this exploratory probe, the best-performing history length shifts from $16$ to $32$, $64$, and $64$\,s as the prediction horizon increases. Gains from additional history saturate earlier for near-term prediction, whereas longer horizons benefit from more context.

\paragraph{For the pre-training objective, the window is not the binding constraint.} At the shortest probe horizon of $1$\,s, gains from longer history largely saturate within $32$\,s. This provides indirect support for our pre-training window length, since the probe predicts a $4$\,s histogram of coarse codes rather than the next per-channel code. Consistent with this finding, extending pre-training sequences from $32$ to $128$\,s did not improve performance on any downstream task. A separate probe that holds the target codes fixed and varies how many past steps a trained backbone may condition on agrees independently: a model pretrained on $128$\,s sequences fails to beat one pretrained on $32$\,s sequences at every history length up to $64$\,s. We do not shorten the sequence below $32$\,s either, for a reason internal to the architecture rather than to this table: at $64$ steps the memory bank is written $16$ times and evicts $8$ times, whereas a $16$\,s sequence would produce exactly $8$ writes and never exercise eviction at all, leaving the model unexposed to memory-bank turnover during pre-training.

\paragraph{For downstream tasks, context length does matter.} The useful context length for downstream tasks is evaluated directly in \S\ref{sec:sequencetask}. Sleep staging is such a task, and the objective we pretrain with is not: the model is trained on a local prediction problem and then asked, at evaluation time, to carry state across intervals far longer than any it saw during pre-training. \S\ref{sec:sequencetask} measures what that is worth, and the fact that it is worth anything is a statement about the recurrent state of \S\ref{sec:backbone}, not about the length of the pre-training window.

\section{Downstream Tasks and Metrics}
\label{app:downstream_task}
We evaluate on five datasets in two families (Table~\ref{tab:tasks}): three \emph{window} tasks, where one segment of a few seconds carries one label, and two \emph{sequence} tasks, where a recording is labelled unit by unit in order: a \emph{unit} being the interval the annotation is defined on, a $30$\,s scored epoch for Sleep-EDF and a $30$\,s interval labelled from the seizure annotations for CHB-MIT. None appears in our pre-training corpus.

\begin{table}[t]
\centering
\caption{\textbf{The five downstream tasks.} The three above the rule are window tasks and the two below are sequence tasks. Unit is the interval one label is defined on, \#Ch.\ the electrodes used and Rate their sampling rate in Hz; $L$ is the number of consecutive units per sample, so $L=1$ is a single labelled segment; $K$ is the readout budget, fixed per dataset and identical across models (Appendix~\ref{app:readout}); Selection is the validation criterion a checkpoint is chosen on, and Split how the data is partitioned. $^\dagger$Not a cross-validation (Appendix~\ref{app:kaggleern}). $^\ddagger$Folds are over the 24 case folders, which come from 23 subjects; see Appendix \ref{app:chbmit}}
\label{tab:tasks}
\begingroup
\fontsize{7}{9}\selectfont
\setlength{\tabcolsep}{2.5pt}
\renewcommand{\arraystretch}{1.08}
\begin{tabularx}{\linewidth}{
    @{}
    >{\raggedright\arraybackslash}X
    >{\raggedright\arraybackslash}p{2.5cm}
    c
    c
    c
    c
    r
    >{\raggedright\arraybackslash}p{1.4cm}
    >{\raggedright\arraybackslash}p{1.8cm}
    @{}
}
\toprule
\textbf{Dataset}
& \textbf{Task}
& \textbf{Unit}
& \textbf{\#Ch.}
& \textbf{Rate}
& $\mathbf{L}$
& $\mathbf{K}$
& \textbf{Selection}
& \textbf{Split} \\
\midrule
FACED \citep{chen2023large}
& Emotion, $9$-class
& $10$\,s
& $30$
& $250$
& $1$
& $19{,}200$
& $\kappa$
& subject, fixed \\

KaggleERN
\citep{inria-bci-challenge,margaux2012objective}
& ERN, binary
& $2$\,s
& $19$
& $200$
& $1$
& $3{,}840$
& ROC-AUC
& subj., $4$ folds$^{\dagger}$ \\

SEED-VIG \citep{1741-2552-14-2-026017}
& Vigilance, regression
& $8$\,s
& $17$
& $200$
& $1$
& $19{,}200$
& $R^2$
& session, fixed \\
\midrule
Sleep-EDF
\citep{kemp2000analysis,goldberger2000physiobank}
& Sleep stage, $5$-class
& $30$\,s
& $2$
& $100$
& $20$
& $12{,}000$
& $\kappa$
& subject, $5$-fold \\

CHB-MIT
\citep{goldberger2000physiobank,shoeb2009application}
& Seizure, binary
& $30$\,s
& $15$
& $256$
& $20$
& $90{,}000$
& AUC-PR
& case, $5$-fold$^\ddagger$ \\
\bottomrule
\end{tabularx}
\endgroup
\end{table}

\subsection{FACED}
Nine-class cross-subject emotion recognition, under the protocol of CBraMod \citep{wang2025cbramod}, on a corpus of $32$-channel recordings from $123$ subjects watching $28$ emotion-eliciting clips \citep{chen2023large}. We use the officially released preprocessing, in which every trial is $30$\,s at $250$\,Hz, and apply no further filtering, artefact rejection or re-referencing. Two of the $32$ channels are mastoid electrodes, HEOR and HEOL, and are discarded, leaving the $30$ EEG channels of the reference protocol; each trial is cut into three non-overlapping $10$\,s segments, giving $84$ segments per subject and $10{,}332$ in total. Subjects are sorted by identifier and partitioned subject-independently into the first $80$ for training ($6{,}720$ segments), the next $20$ for validation ($1{,}680$) and the last $23$ for test ($1{,}932$); the split is deterministic and identical for every model and seed. Class proportions are fixed by the stimulus design rather than by sampling and are identical across the three splits. We report balanced accuracy and Cohen's $\kappa$ over five seeds. Balanced accuracy averages per-class recall, so any input-independent predictor scores exactly $1/9$; plain accuracy would instead depend on which degenerate strategy a collapsed model settled into. Two cautions about the test set: with the split fixed, seeds vary only initialisation and batch order, so the reported dispersion is optimisation variance and not variance across subjects; and the $1{,}932$ test segments come from $644$ subject-video trials, $23$ subjects by $28$ videos, so the three segments of one trial should not be treated as independent.

\subsection{KaggleERN}
\label{app:kaggleern}
Binary detection of error-related negativity, under the protocol of EEGPT \citep{wang2024eegpt}, on the corpus of the BCI Challenge @ NER 2015, in which each item selection in a P300 speller is followed by a visual feedback the subject may perceive as erroneous \citep{inria-bci-challenge,margaux2012objective}. The release holds $56$-channel EEG from $26$ subjects at $200$\,Hz, $340$ feedback-locked trials each, $8{,}840$ in total, with the competition's own partition into $16$ labelled subjects ($5{,}440$ trials) and $10$ held-out test subjects ($3{,}400$); the classes are imbalanced at $70.8\%$ and $70.9\%$ positive respectively. Trials are epoched from $-0.7$ to $+1.3$\,s and we keep the $19$-channel standard $10$--$20$ subset of the reference protocol, with no band-pass, baseline correction or re-referencing, again following it.

Two departures are worth stating. The four folds are not a cross-validation: all four score on the same fixed ten test subjects and differ only in which group of four labelled subjects is removed from training, so their dispersion measures sensitivity to the training subset and not variance across test subjects. Our four training sets match the published fold dictionary exactly, and the seed is fixed so that the dispersion is not confounded with initialisation. Second, the reference uses the ten test subjects as its validation set, which combined with checkpoint selection on a validation criterion would make every reported number a best-epoch-on-the-reported-set. We instead hold out $15\%$ of the training subjects' trials, stratified by subject and label, giving roughly $3{,}468/612/3{,}400$ trials per fold, with the split keyed on the fold index so that all models see the same partition. We report ROC-AUC, the metric of the competition and of the reference protocol, alongside AUC-PR.

Ground-truth labels for the ten test subjects are not distributed by the competition; we use the publicly circulated label file that the reference implementation instructs the user to supply, and report two consistency checks on it rather than an independent verification of its provenance. Its positive rate is $0.7091$ against $0.7077$ in the released training labels, and scoring the competition winner's two published submissions against it yields ROC-AUC $0.8679$ and $0.8439$, where a label file unrelated to the data would score near $0.5$. These are consistency checks and not an independent confirmation of where the file came from.

\subsection{SEED-VIG}
\label{app:seedvig}
Continuous vigilance regression, under the protocol of CBraMod \citep{wang2025cbramod}, from $17$-channel EEG at $200$\,Hz recorded during simulated driving, with a PERCLOS value in $[0,1]$ derived from eye tracking once per $8$\,s window \citep{1741-2552-14-2-026017}. The corpus holds $23$ sessions from $21$ subjects (two subjects contributed two sessions) of $885$ windows each, $20{,}355$ in total, and the target is a scalar per window rather than a sequence. We follow the reference protocol, ordering sessions by numeric index and taking the first $15$ for training, the next $4$ for validation and the last $4$ for test ($13{,}275/3{,}540/3{,}540$); every subject falls entirely within one split. We report Pearson correlation and $R^2$ over five seeds, and select on validation $R^2$.

The session ordering is consequential, and the reference is not self-consistent about it. CBraMod's text describes the split as ``subject 1 to 15 for training, subject 16 to 19 for validation and subject 20 to 23 for test'' \citep{wang2025cbramod}, which can only be read as file indices in numeric order, there being $23$ sessions from $21$ subjects. Its released preprocessing code instead sorts filenames lexicographically, and because those filenames are not zero-padded, sessions \texttt{10}--\texttt{19} sort ahead of session \texttt{2}; the first fifteen files are then $1,10,\dots,19,2,20,21,22$ rather than $1,\dots,15$, which moves both ends of every split. A reproducer following the text and a reproducer running the code do not evaluate on the same data.

We resolved this from published results rather than by choosing. Because $R^2=1-\mathrm{MSE}/\mathrm{Var}$, any paper reporting both $R^2$ and RMSE pins the variance of its own test labels at $\mathrm{Var}=\mathrm{RMSE}^2/(1-R^2)$. Four published rows, three from CBraMod's own comparison table \citep{wang2025cbramod} and one from a later model \citep{zhou2026csbrain}, imply a test-label standard deviation of $0.3175$--$0.3176$. We measure $0.3176$ under numeric ordering and $0.175$ under lexicographic. The published numbers therefore follow the text and not the released code, and we follow the published numbers.

The difference is not a detail. Under lexicographic ordering three of the four test sessions are almost flat, so $R^2$ is measured against a very small denominator: predicting the training mean scores $-0.413$ there, against $-0.019$ under numeric ordering. A reproducer who ran the released code would obtain implausibly poor numbers for every model, ours included, with nothing in the output to indicate why.

An $R^2$ on this task is easy to misread, so we give three reference points. A predictor that always emits the training mean scores $R^2=-0.019$, at RMSE $0.3206$. One that cheats and emits the \emph{test} mean scores $R^2=0$ by construction, at RMSE $0.3176$. The best published methods reach $R^2\approx0.24$--$0.34$, at RMSE $\approx0.26$. The entire competitive range is thus an RMSE from $0.32$ down to $0.26$: absolute errors move very little, and an $R^2$ of a third is a strong result here rather than a weak one.

Two further properties are worth stating. The obvious shortcut is closed: $90.2\%$ of the test labels' variance lies \emph{within} a session and only $9.8\%$ between sessions, so a model cannot score by learning how drowsy each driver is on average, it has to follow the change over minutes. And a regression head has a degenerate solution, a constant output, which yields $R^2\approx0$ and an undefined correlation while looking from the outside like ordinary underperformance; we checked for it, and none of the thirty runs collapsed, the lowest single-run correlation being $0.454$.

\subsection{Sleep-EDF}
\label{app:sleepedf}
Five-class sleep staging on the Sleep Cassette subset \citep{kemp2000analysis, goldberger2000physiobank}, preprocessed as in EEGPT \citep{wang2024eegpt}, $153$ whole-night polysomnograms from $78$ subjects, of which we use only the two EEG derivations, Fpz--Cz and Pz--Oz at $100$\,Hz. Recordings are ambulatory and run for a median of $23$ hours, so most of the raw signal is daytime wakefulness; following the convention of the field we retain $30$ minutes of wake either side of the first and last non-wake epoch. Annotations are mapped to the five AASM classes with stages 3 and 4 merged, and movement and unscored epochs discarded, leaving $195{,}469$ epochs of $30$\,s, which reproduces the standard $78$-subject configuration to within ten epochs and matches its per-class counts for N1, N2, N3 and REM exactly. Sequences are cut with stride $L$ so that no sequence crosses a night and every epoch is predicted exactly once; a recording whose epoch count is not a multiple of $L$ yields one short final sequence, zero-padded, with padded positions carrying an ignore label.

The two EEG signals are bipolar derivations, whereas four backbones address channels by electrode identity and no lookup table contains an entry for a derivation pair. We name each channel by the leading electrode of its pair, Fpz--Cz $\to$ Fpz and Pz--Oz $\to$ Pz, and verified that both names resolve in every table before training; this costs no parameters and adds no layer. We considered and rejected the alternative used by the reference sleep pipeline, a learned $1\times1$ convolution from two channels to the backbone's channel count, because that places a randomly initialised layer in front of every pre-trained backbone, precisely the configuration that drives a constant-output solution (Appendix~\ref{app:headact}), and with two input channels the mixing matrix has rank at most two, so the risk is higher here rather than lower. BIOT retains its own learned channel mixing, which is intrinsic to that model on every dataset because its pre-trained channel tokens describe montage pairs rather than electrodes.

Together with CHB-MIT, this is one of the two datasets whose partition is not fixed by a reference protocol, and therefore one of the two that report cross-subject rather than optimisation variance. We use subject-wise five-fold cross-validation: the $78$ subjects are permuted once with a fixed seed and split into groups of $16/16/16/15/15$; fold $k$ tests on group $k$, validates on $10\%$ of the remaining subjects drawn with a fold-derived seed so that all models see the same partition, and trains on the rest. Both nights of a subject always move together. We report macro-F1 and $\kappa$ as the mean and standard deviation of the five per-fold test scores, and select on validation $\kappa$, as on FACED.

\subsection{CHB-MIT}
\label{app:chbmit}
Binary seizure detection on $980$ hours of continuous paediatric scalp EEG from $23$ subjects across $24$ case folders, distributed as $683$ segments of roughly one hour \citep{goldberger2000physiobank, shoeb2009application}. Four properties of the release change the resulting benchmark and none of them raises an error, so a pipeline that gets them wrong produces a plausible dataset rather than a failure.

\paragraph{Seizure annotations appear in two formats.} The summary files record onsets either as \texttt{Seizure Start Time:} or as \texttt{Seizure $n$ Start Time:}, and which form is used varies by subject rather than by how many events a file contains: \texttt{chb08} writes \texttt{Seizure 1 Start Time:} for files holding a single seizure, while \texttt{chb01} and \texttt{chb24} never number theirs. A parser matching only the unnumbered form recovers $40$ of the $198$ annotated seizures and reports no error. The remaining $158$ are not dropped but silently relabelled interictal, so the model is trained on seizure activity presented as normal, and the positive rate falls from $0.456\%$ to below one in a thousand.

\paragraph{Amplitudes are stored in volts.} Read without conversion the signals are six orders of magnitude below the microvolt-scaled data of every other corpus we use, and nothing in the pipeline would raise an error on it.

\paragraph{Channel names collide.} The segments carry three incompatible bipolar montages. Backbone channel lookups are keyed by one electrode name, so keying on the first electrode of each derivation collides \texttt{FP1-F3} with \texttt{FP1-F7} and \texttt{FP2-F4} with \texttt{FP2-F8}; we resolve each in favour of the temporal chain, leaving $15$ distinct electrodes. Separately, three \texttt{chb12} segments holding $13$ seizures are recorded against a common reference rather than in the bipolar chain used elsewhere and are excluded, leaving $683$ of $686$ segments and $185$ of $198$ seizures ($93.4\%$), so that a difference in signal type is not confounded with the class. Preparation yields $117{,}553$ units of $30$\,s over $15$ channels, of which $0.456\%$ contain seizure activity.

\paragraph{The validation split needs four subjects, not two.} Splits are five-fold over case folders, so every unit is tested exactly once. At a positive rate below one percent the usual ten-percent validation rule is too small to select on: two validation subjects give folds holding as few as $28$ positive units, and on exactly those folds two different baseline architectures both fall below the random AUC-PR baseline of $0.0046$, the same folds for both models, which identifies the split rather than the models. We instead allocate four, chosen by enumerating all four-subject combinations and taking the one whose share of positives is closest to twenty percent, with ties broken in favour of subjects not yet used for validation; validation positives rise to $77$--$91$.

\paragraph{Two case folders are one subject.} The release contains $24$ case folders from $23$ subjects: \texttt{chb21} was recorded from the same person as \texttt{chb01}, a year and a half later. Our folds are drawn over case folders, so that subject is treated as two, and in two of the five folds one of its folders is in the test set while the other is in training; every other subject appears on one side only. We measured what this is worth by removing the shared subject's test units from those two folds and recomputing the AUC-PR of every row of Table~\ref{tab:long_sequence_results}. It moves by at most $0.018$ on the fifteen-channel montage and by at most $0.009$ on the two-channel one: NeurDuo (Small) from $0.413$ to $0.409$, Base from $0.392$ to $0.384$, Large from $0.471$ to $0.476$, the strongest classical pipeline from $0.285$ to $0.284$ and the strongest pre-trained baseline from $0.208$ to $0.217$. Every NeurDuo variant retains its AUC-PR advantage over every baseline on both montages.

\paragraph{AUROC alone is not safe here.} Segments are long and seizures cluster within them, so knowing \emph{which recording} a window came from is by itself predictive of its label. An oracle that never sees the signal and is told only whether a unit's recording contains a seizure reaches AUROC $0.905$ averaged over our five test folds but AUC-PR only $0.024$. Recording-level label structure therefore settles most of the AUROC and almost none of the precision--recall area. It is not a baseline: it uses information no model receives, and what it measures is how much of each metric is decided by which recording a unit belongs to rather than by the signal in it. We therefore select on validation AUC-PR and report AUC-PR and false alarms per hour, with the random AUC-PR baseline of $0.0046$ alongside.

\paragraph{How a false alarm is counted.} For each evaluation fold, the threshold is the median score of the positive units, taken halfway between the two central scores when their number is even, so that at least $50\%$ of positive units, not seizure events, reach it. The units of a fold are scored as one sequence, recording after recording, and consecutive predicted-positive units are merged into one alarm, which counts as false if it overlaps no seizure-labelled unit; predicted-positive units on either side of the boundary between two consecutive recordings therefore form a single alarm. FP$\cdot$h$^{-1}$ is computed by dividing the total false alarms by the total recording hours across all five folds. Because thresholds are selected on the evaluated folds rather than on validation data, this metric compares models at a common sensitivity target rather than measuring deployment performance with a fixed threshold.

\paragraph{Comparison with prior work.} \citet{CaMBRAIN} is the closest prior work, a causal state-space model for streaming EEG with its own persistent-state ablation on this corpus, and we take its metrics and its choice not to filter recordings. We do not take its split, which is file-level random and so places recordings of one patient on both sides, an arrangement that, given a shortcut worth AUROC $0.905$, is considerably easier than our folds, which keep whole recordings and, with the one exception noted above, whole subjects on one side. Its window is $5$\,s against our $600$. The published numbers above are quoted as the scale of the benchmark, not as a comparison.

\section{Baselines and Evaluation Protocol}
\label{app:baselinesandevaluation}
\subsection{Baselines, and what each saw before it saw the task}
\label{app:overlap}
Table~\ref{tab:baselines} lists the nine baselines of \S\ref{sec:downstream}. The six pre-trained models span the design space of this class deliberately: backbones from $3.2$ to $69.2$ million parameters, pre-training corpora from a single clinical archive to $92$ aggregated datasets, and tokenizers dividing the signal at granularities from $0.125$ to $1$\,s per channel. The three classical pipelines span the three feature families classical EEG decoding exploits and are not head-matched, a classical pipeline having no separable backbone and readout.

\begin{table}[!t]
\centering
\caption{\textbf{The nine baselines.} The six above the rule are pre-trained foundation models and the three below are decoders that were never pre-trained. \#Params is the pre-trained backbone as loaded, excluding pre-training-only components such as decoders and projection heads; a classical pipeline has no separable backbone and readout, so the column is left blank. Pre-training reproduces each release's own account of its corpus, which is why the entries are not stated on a common footing.}
\label{tab:baselines}
\begingroup
\fontsize{7}{9}\selectfont
\setlength{\tabcolsep}{2.5pt}
\renewcommand{\arraystretch}{1.08}
\begin{tabular}{@{}lp{4.6cm}rp{3.6cm}@{}}
\toprule
\textbf{Baseline}
& \textbf{Architecture / feature family}
& \textbf{\#Params}
& \textbf{Pre-training} \\
\midrule
BIOT
& Linear-attention transformer over per-channel STFT patches
& $3.2$M
& MGH, SHHS, TUAB, TUEV, CHB-MIT, IIIC \\

CBraMod
& Criss-cross transformer over the channel $\times$ time grid
& $4.9$M
& TUEG, ${>}9{,}000$\,h \\

LaBraM-Base
& Vector-quantised tokenizer with masked-token prediction
& $5.8$M
& $16$ corpora, ${>}2{,}500$\,h \\

EEGPT-Large
& ViT with dual mask-reconstruction and alignment
& $25.3$M
& PhysioNet-MI, TSUBenchmark, M3CV, SEED \\

ST-EEGFormer-S
& ViT with masked autoencoding on raw EEG
& $25.4$M
& ${>}8$M segments; MI, P300, SSVEP \\

REVE-Base
& ViT with 4D Fourier encoding of electrode geometry
& $69.2$M
& $92$ datasets, ${>}60{,}000$\,h, ${\sim}25{,}000$ subjects \\
\midrule
BP-GBDT
& Spectral power: band power, differential entropy and shape statistics $\to$ gradient boosting
& ---
& none \\

FBCov-TS-Lin
& Spatial covariance: filter-bank covariances $\to$ Riemannian tangent space $\to$ linear
& ---
& none \\

ERP-Lin$^\dagger$
& Evoked waveform: band-limited, decimated, standardised windows $\to$ linear
& ---
& none \\
\bottomrule
\end{tabular}
\endgroup

\vspace{8pt}
\parbox{\linewidth}{\footnotesize
$^\dagger$Adapted from the evoked-response pipeline of \citet{yang2026eeg} by dropping its supervised spatial filter, which requires discrete classes, so that one recipe spans our classification, regression and sequence-labelling tasks; the band-pass is unchanged, and so is the decimation rate except on Sleep-EDF as described above.}
\end{table}

\subsubsection{The classical pipelines}
These three exist to answer a question the shared head cannot: whether any pre-trained model beats a decoder with no pre-training at all. They are deliberately not head-matched, since a classical pipeline has no separable backbone and readout, but everything defining the task is matched. They read the same split files by the same rules, including the fold-derived stratified validation draw of KaggleERN, so the validation trials are literally the same ones. Their metric functions are the same \texttt{sklearn} calls as the neural evaluator, key for key. And their selection criterion is the same: where a fine-tuned model selects an epoch, each of the two linear pipelines selects one regularisation scalar on validation from a log-spaced ladder ($C\in[10^{-8},10^{3}]$ for classification, $\alpha\in[10^{-2},10^{10}]$ for regression).

\paragraph{BP-GBDT (spectral power).} Welch spectra give twelve descriptors per channel: differential entropy in five bands ($1$--$4$, $4$--$8$, $8$--$13$, $13$--$30$, $30$--$45$\,Hz), log broadband power, log standard deviation, skewness, kurtosis, Hjorth mobility and complexity, and spectral entropy scaled to $[0,1]$. Differential entropy under a Gaussian is one half of the logarithm of band power up to an additive constant, the standard feature of the emotion and vigilance literature; the statistical and entropy terms are what a sleep-staging feature set needs \citep{vallat2021open,kastrati2025eeg}. Features go to LightGBM at $300$ trees and learning rate $0.05$. This pipeline has no grid, so it cannot benefit from validation selection at all.

\paragraph{FBCov-TS-Lin (spatial covariance).} Band-pass into the same five bands, an OAS-shrunk covariance matrix per band, projection to the Riemannian tangent space, and a linear model \citep{jayaram2018moabb,sabbagh2019manifold}.

\paragraph{ERP-Lin (evoked waveform).} Band-pass to $1$--$20$\,Hz, decimate to $32$\,Hz, per-channel per-window standardisation, flatten, $\ell_2$-regularised linear model. One dataset forces a deviation, recorded in the result files rather than applied silently: Sleep-EDF decimates to $16$\,Hz because the context rule below triples its feature vector.

\paragraph{Sequence tasks.} On the sequence tasks a classical pipeline has no recurrence, so we give it context explicitly: each unit's feature vector is concatenated with the mean of the ten units before it and the mean of the ten after, within the same recording. This gives the classical pipelines bidirectional context, an advantage over the causal context head of the neural models; units near the edges of a recording use their own features in place of a missing side.

\subsubsection{Pre-training overlap}
Interpreting transfer performance requires accounting for each model’s pre-training data. We audited the pre-training corpus of every model, our own included, against every downstream dataset. Table~\ref{tab:overlap} summarises; the paragraphs below give the cases that are not simply clean. We distinguish two kinds of contamination, because they are not equally serious. \emph{Corpus overlap} means that a downstream corpus is included in a model's pre-training data. Whether that includes the labels, and which of the recordings we test on it covers, depends on the released checkpoint and on the split its authors used, so we state both per case below. \emph{Domain overlap} means the pre-training data contains a different corpus of the same kind (the same recording paradigm, montage class and clinical population) which confers an advantage that is real but is the ordinary business of pre-training rather than leakage.

\begin{table}[!t]
\centering
\caption{\textbf{Pre-training overlap by model and task.} Every model's pre-training corpus, ours included, was audited against every downstream dataset. $\bullet$ marks corpus overlap, the downstream corpus itself appearing in that model's pre-training set; it occurs twice and both cases are discussed below. $\circ$ marks domain overlap, a different corpus of the same recording paradigm and population, which is the ordinary condition of pre-training at scale rather than leakage. Blank means neither. The three classical pipelines of Table~\ref{tab:baselines} have no row here, never having been pre-trained.}
\label{tab:overlap}
\begingroup
\fontsize{7}{9}\selectfont
\setlength{\tabcolsep}{2.5pt}
\renewcommand{\arraystretch}{1.08}
\begin{tabular}{@{}lccccc@{}}
\toprule
\textbf{Model}
& \textbf{FACED}
& \textbf{KaggleERN}
& \textbf{SEED-VIG}
& \textbf{Sleep-EDF}
& \textbf{CHB-MIT} \\
\midrule
BIOT
& & & & $\circ$ & $\bullet$ \\
CBraMod
& & & & & $\circ$ \\
LaBraM-Base
& $\circ$ & $\bullet$ & & & $\circ$ \\
EEGPT-Large
& $\circ$ & & & & \\
ST-EEGFormer-S
& & & & & \\
REVE-Base
& & & & $\circ$ & $\circ$ \\
\midrule
\textbf{NeurDuo (ours)}
& & & & $\circ$ & $\circ$ \\
\bottomrule
\end{tabular}
\endgroup
\end{table}

\paragraph{KaggleERN and LaBraM: corpus overlap.} LaBraM's pre-training set lists the Inria BCI Challenge as the fourth of its sixteen corpora: KaggleERN. This is not domain similarity but the same corpus, and since the competition releases the raw EEG of all twenty-six subjects while LaBraM's pre-training is unlabelled self-supervision, its encoder has most likely seen the signals of our ten test subjects, though never their labels. No overlap-free LaBraM checkpoint exists; we mark the row and do not exclude it, since removing the model would remove a reference point the rest of the field uses.

\paragraph{CHB-MIT and BIOT: corpus overlap.} We load the released 'EEG-six-datasets-18-channels' checkpoint, which its authors describe as pre-trained on resting and sleep corpora together with the training sets of TUAB, TUEV, IIIC and CHB-MIT. The CHB-MIT part is supervised: in the released pre-training script that corpus enters the loss through its seizure labels, and its split is by case folder, with \texttt{chb01} to \texttt{chb20} in training. Twenty of the twenty-four case folders we evaluate on are therefore folders this checkpoint was trained on with labels, covering $88.6\%$ of our test units and $87.1\%$ of our test positives. We mark the row rather than drop it, and we keep the same checkpoint across every task so that the BIOT of this table is the BIOT of the other four; we do not substitute the overlap-free \texttt{EEG-PREST} checkpoint for the same reason. What the row should be read as is a baseline that was given labelled seizure supervision on most of the subjects it is scored on here.

\subsection{Downstream protocol}
\label{app:downstream}
Published protocols are reproduced here where they exist rather than invented, and we name them once so that the shorthand below is unambiguous. FACED and SEED-VIG follow CBraMod~\citep{wang2025cbramod}, and KaggleERN follows EEGPT~\citep{wang2024eegpt}; the per-model input conventions and the warm-up schedule follow \citet{yang2026eeg}. Sleep-EDF takes its wake cropping from EEGPT but uses our own subject-wise split (Appendix~\ref{app:sleepedf}), and CHB-MIT takes its choice of recordings and metrics from CaMBRAIN~\citep{CaMBRAIN} but uses our own split and selection (Appendix~\ref{app:chbmit}). Where a paragraph says \emph{the reference}, it means whichever of these protocols governs the dataset under discussion. Deviations from a reference are stated where they occur and are applied identically to every model.

\subsubsection{The shared readout}
\label{app:readout}
\paragraph{Structure.} A backbone exposes $\mathrm{tokens}(x)\in\mathbb{R}^{B\times N\times D}$; everything after that point is common to all models. The head is $\mathrm{LN}$, a per-token linear map $D\to d_{\mathrm{tok}}$ with no activation, a flatten of the whole grid to $N d_{\mathrm{tok}}$, then $\mathrm{Linear}\to\mathrm{LeakyReLU}(0.01)\to\mathrm{Linear}$ with a hidden width of $256$ and dropout of $0.1$ before each of the two linear layers. The grid is flattened rather than pooled: pooling to one vector per window is the capacity bottleneck the shared head exists to remove, and reintroducing it would restore the confound. For a fixed budget $K$, tasks differ only in the width of the final layer, which moves at most $2{,}056$ parameters. The $\mathrm{LN}$ is best read as normalisation of the backbone interface rather than as head capacity, it costs $2D$ parameters, about $0.02\%$ of the head, and it is not optional: measured on one FACED batch the tokens arriving at the head have standard deviations spanning more than an order of magnitude across backbones, one of them uncentred at mean $+1.5$, so without it a single head learning rate would correspond to very different effective rates per model. The choice of LeakyReLU over the more usual rectifiers is also forced, for a reason given in Appendix~\ref{app:headact}.

\paragraph{How $K$ is set.} Fixing $d_{\mathrm{tok}}$ would make head size track token count and reward whichever tokenizer emits the most tokens; we therefore fix $K=N\,d_{\mathrm{tok}}$ and let $d_{\mathrm{tok}}=\mathrm{round}(K/N)$ vary. The per-token width then differs across models: $8$ dimensions for ST-EEGFormer against $120$ for EEGPT on FACED. But a token is not a comparable unit across models: an ST-EEGFormer token covers $0.125$\,s of one channel and a CBraMod token covers $1$\,s of one channel, so those are the same capacity per channel-second. The constraint that actually determines $K$ on every dataset is that no model's $d_{\mathrm{tok}}$ may exceed its width $D$ and be silently clamped, since a clamp reintroduces exactly the spread the budget removes. Where that constraint is slack the value comes from holding readout capacity per second of EEG fixed at the $1{,}920$ dimensions per second set a priori on FACED. Where it binds, $K=\min_m N_m D_m$. Table~\ref{tab:budget} gives the resulting values and what they buy. Two consequences are worth naming. On KaggleERN the constraint is not hypothetical: at $19{,}200$ the $2$\,s window leaves $N=32$--$57$ tokens for four of the six baselines, their $d_{\mathrm{tok}}$ clamps, and head sizes spread by a factor of $2.56$ ($1.99$--$5.09$M) rather than $1.045$. And on SEED-VIG the FACED value was carried over unchanged to an $8$\,s window, so that task runs at $2{,}400$ dimensions per second rather than $1{,}920$; the constraint is slack there and nothing clamps, but we record the inconsistency rather than present a rule the numbers do not follow.

\begin{table}[!htbp]
\centering
\caption{\textbf{Readout budget per dataset and the head sizes it produces.} $K=N d_{\mathrm{tok}}$ is the product of token count and per-token width, held fixed per dataset so that head size does not track whichever tokenizer emits the most tokens. Determined by names the rule that sets it: either a fixed readout capacity per second of EEG, or the no-clamp bound $\min_m N_m D_m$ where holding that capacity would push some model's $d_{\mathrm{tok}}$ above its width $D_m$. Head size is the range over the nine models that receive a shared head and Spread their ratio; the Sleep-EDF and CHB-MIT figures include the context head, which is identical across models and therefore compresses the ratio.}
\label{tab:budget}
\begingroup
\fontsize{7}{9}\selectfont
\setlength{\tabcolsep}{2.5pt}
\renewcommand{\arraystretch}{1.08}
\begin{tabular}{@{}lrlrl@{}}
\toprule
\textbf{Dataset}
& $\mathbf{K}$
& \textbf{Determined by}
& \textbf{Head size}
& \textbf{Spread} \\
\midrule
FACED
& $19{,}200$
& $1{,}920$ dim/s at $10$\,s
& $4.92$--$4.98$M
& $1.012\times$ \\
KaggleERN
& $3{,}840$
& $1{,}920$ dim/s at $2$\,s
& $1.00$--$1.05$M
& $1.045\times$ \\
SEED-VIG
& $19{,}200$
& carried over from FACED
& $4.93$--$5.02$M
& $1.020\times$ \\
Sleep-EDF
& $12{,}000$
& no-clamp bound $\min_m N_m D_m$
& $4.75$--$4.94$M
& $1.041\times$ \\
CHB-MIT (15 Ch)
& $90{,}000$
& no-clamp bound $\min_m N_m D_m$
& $24.81$--$24.95$M
& $1.006\times$ \\
CHB-MIT (2 Ch)
& $12{,}000$
& no-clamp bound $\min_m N_m D_m$
& $4.75$--$4.94$M
& $1.041\times$ \\
\bottomrule
\end{tabular}
\endgroup
\end{table}

\subsubsection{The context head}
A sequence sample is $L$ consecutive \emph{units} of one recording, a unit being the interval one label is defined on: a $30$\,s scored epoch for Sleep-EDF and a $30$\,s interval labelled from the seizure annotations for CHB-MIT. We say \emph{unit} rather than \emph{window} because a window in this paper is either the $32$\,s pre-training sequence of \S\ref{sec:pretrain} or the single labelled segment of a window task, and rather than \emph{epoch} because a training epoch is a pass over the corpus. Each unit is encoded using the same backbone weights. The baselines process units independently, whereas NeurDuo carries its recurrent state across unit boundaries. That state is not detached at the boundaries during fine-tuning, so the backbone is trained through them and gradients reach earlier units. A shared head then converts each unit’s token grid into a 256-dimensional representation, using its hidden activation as the output. Finally, a context head takes the resulting $(B,L,256)$ sequence and predicts a label for each unit. Using the shared head's own hidden layer rather than a pooled vector is what carries the flatten-the-grid and equal-$K$ properties over to the sequence tasks unchanged.

The context head we report is two Mamba blocks ($d_{\mathrm{model}}=256$, $d_{\mathrm{state}}=16$, convolution width $4$, expansion $4$) followed by layer normalisation and a linear map at every position, with padded positions zeroed at the input and excluded from the loss. It has $1{,}753{,}861$ parameters and is byte-identical across all models. It is causal, and it carries no positional embedding, the recurrence being ordered by construction. We report this setting rather than the bidirectional one because a bidirectional readout recomputes over all $L$ units whenever a unit arrives, which surrenders in the readout the streaming property the backbone was built to have; a comparison that allowed it would not be measuring what the paper claims.

To assess the sensitivity of NeurDuo to the choice of context head, we also evaluated a bidirectional alternative: two Transformer encoder layers, $8$ heads, feed-forward width $1{,}024$, a learned positional table of length $L$, $1{,}586{,}437$ parameters on our own model, holding everything else fixed. It is parameter-matched to within $11\%$, deliberately: an unmatched comparison would confound the cost of causality with the cost of capacity. On Sleep-EDF at $L=20$ with state carried, the causal head reaches macro-F1 $0.7553\pm0.0081$ and the bidirectional head $0.7532\pm0.0150$ over the same five folds. Sleep staging is a task where future context is genuinely informative and the standard sequence models for it are bidirectional, so we expected a loss and did not find one.

\subsubsection{Input conventions}
These models were pre-trained at different sampling rates and in different amplitude units, so feeding all of them one identical tensor would place several out of distribution and would measure that rather than the quality of the representation. Each therefore receives the signal resampled and rescaled to its own pre-training convention, following the reference implementation and the model's own release otherwise; the one exception is the amplitude scaling on KaggleERN, where the reference protocol prescribes a single convention that we apply to every model alike (Appendix~\ref{app:kaggleern}). What is held identical is everything the task consists of: the same channels, the same segments, the same labels and the same splits; only the input representation differs, and all pre-trained weights are loaded in full for every model.

\subsubsection{Optimisation}
\label{app:optimisation}
All models are fine-tuned end to end, every parameter, under one recipe: AdamW with weight decay $0.05$; backbone learning rate $10^{-4}$, not batch-size scaled; head learning rate $10^{-3}\sqrt{b/256}$ at batch $b$, the square-root rule of CBraMod's released fine-tuning code \citep{wang2025cbramod}, which is $5\times10^{-4}$ at $b=64$ and $3.54\times10^{-4}$ at $b=32$; three linear warm-up epochs \citep{yang2026eeg} then a per-step cosine schedule annealed to $10^{-6}$; gradient-norm clipping at $1.0$. Multi-class and sequence-labelling tasks use cross-entropy with label smoothing 0.1, the binary KaggleERN task binary cross-entropy, and regression mean squared error. We do not transplant each model's published recipe, since those were tuned on the datasets of their own papers.

Two models force a deviation, and both are handled by the recipe rather than by a separate decision. ST-EEGFormer runs at batch $32$ because its $2{,}400$-token sequence does not fit at $64$, and at $80$ rather than $160$ units per step on the sequence tasks; its head learning rate follows from the same rule at the smaller batch, and we halve its batch on \emph{every} dataset, including those where the full batch would fit, so that its batch stands in the same relation to the common recipe on every dataset rather than only where memory forces the deviation. LaBraM's temporal embedding table has $16$ rows at one second per patch and therefore cannot encode a $30$\,s epoch in one pass; rather than truncate or pool the epoch we split it into two $15$\,s sub-windows, encode each, and concatenate the tokens, which leaves the token count, the budget and the head size unchanged and discards no sample. What is lost is self-attention across that internal boundary. LaBraM alone pays it, it is forced by the size of a pre-trained table rather than chosen, and we name it because it is a genuine asymmetry.

The window tasks train for $50$ epochs at batch $64$. The sequence tasks budget the batch in \emph{units} rather than in sequences: the loader takes $\max(1,\mathrm{round}(160/L))$ sequences, which is $160$ units per optimiser step at every $L$ that divides it, $180$ at $L=60$ and $120$ at $L=120$, the two lengths where no integer number of sequences hits the budget. We budget units rather than sequences because fixing the number of sequences instead would let the schedule move with $L$: the data seen per step would scale as $L$, the optimiser steps per epoch as $1/L$, and the head learning rate, which follows $\sqrt{b/256}$ in the batch $b$, as $\sqrt{L}$. The context-length results would then be confounded with all three. The sequence tasks hold many more samples than the window tasks, Sleep-EDF about twenty times as many as FACED, so their number of passes is set by the gradient steps it produces rather than copied across: Sleep-EDF and CHB-MIT train for $20$ epochs, which gives $17{,}400$--$18{,}240$ and $7{,}840$--$10{,}300$ optimiser steps per fold, against $2{,}750$, $5{,}250$ and $10{,}400$ for KaggleERN, FACED and SEED-VIG at $50$ epochs. Copying the window tasks' $50$ epochs would instead have given Sleep-EDF more than $43{,}000$ steps per fold. The remaining difference is not removed, and we note it rather than describe the budgets as matched. ST-EEGFormer, at half the batch, takes twice as many steps throughout.

\subsection{The hidden activation of the shared head}
\label{app:headact}
The two-layer MLP of the shared readout uses LeakyReLU$(0.01)$, and this choice is motivated by the activation comparison below. BIOT is the only backbone in the comparison with a randomly initialised layer (its $30\to18$ channel mixing) in \emph{front} of pre-trained weights. For the first epoch that layer passes almost no signal, so the cheapest available descent direction is to stop reading the input and emit the training class prior, and the cheapest way to realise a constant output is to drive all $256$ hidden units negative. Under an activation whose gradient vanishes on the negative side the run cannot then recover: no gradient reaches the channel mixing, its features never improve, and the loss sits exactly at the class-prior value. We measured this on CPU: $9$ of $256$ units are dead at initialisation, $180$ after fifteen steps and $210$ after thirty, after which the loss holds at $2.19341$ (the class-prior loss to five decimals) for forty-eight consecutive epochs.

We vary only the shared head's hidden activation and train each setting with five seeds, with warm-up disabled so that the activation is the only safeguard. LeakyReLU is the only tested activation for which all five runs escape the class-prior solution (Table~\ref{tab:headact}).

\begin{table}[!htbp]
\centering
\caption{\textbf{Escape from the class-prior solution on FACED with the BIOT backbone.} Five seeds per activation, warm-up disabled, everything else held fixed. A run that fails to escape drives all $256$ hidden units of the shared head negative and emits the training class prior thereafter, at which point balanced accuracy is exactly $1/9$, the chance level for nine classes. Rows are ordered by how much gradient the activation leaves on the negative side, evaluated at $x=-8$, a value typical of the collapsed units. The recipe of the main tables adds three warm-up epochs, which is why the LeakyReLU row reads $0.354$ here against the $0.364$ of Table \ref{tab:window}.}
\label{tab:headact}
\begingroup
\fontsize{7}{9}\selectfont
\setlength{\tabcolsep}{2.5pt}
\renewcommand{\arraystretch}{1.08}
\begin{tabular}{@{}lrrl@{}}
\toprule
\textbf{Activation}
& \textbf{\boldmath$\mathrm{d}/\mathrm{d}x$ at $x=-8$}
& \textbf{Seeds escaping}
& \textbf{Balanced accuracy} \\
\midrule
GELU
& $0$ (and ${<}0$ for $x<-0.75$)
& $0/5$
& $0.111 \pm 0.000$ \\
ReLU
& $0$
& $1/5$
& $0.165 \pm 0.108$ \\
ELU
& $3.4\times10^{-4}$
& $3/5$
& $0.219 \pm 0.124$ \\
LeakyReLU$(0.01)$
& $1.0\times10^{-2}$
& $5/5$
& $0.354 \pm 0.013$ \\
\bottomrule
\end{tabular}
\endgroup
\end{table}

Two points are worth stating because they are not what one would guess. GELU is the \emph{worst} of the four rather than a safe middle ground: for strongly negative inputs it provides almost no gradient, and none of its five runs recovered. And ELU is not sufficient at this width: an earlier head with $d_{\mathrm{tok}}=32$ survived $5/5$ with ELU, but at the $d_{\mathrm{tok}}$ this budget gives, only $3/5$. Uniform warm-up independently rescues $5/5$ and is applied on top, but warm-up lowers the probability of entering the trap, whereas LeakyReLU avoided this failure in all five tested runs. We record this because without warm-up, replacing LeakyReLU with ReLU substantially reduced performance in this diagnostic setting, and because it is the failure mode that decides, in Appendix~\ref{app:sleepedf}, against learning a $1\times1$ channel projection in front of every backbone.

\section{Additional Experiments and Analysis}
\subsection{The Contribution of Pre-training}
\label{app:no_pretrain}

\begin{figure}[!t]
  \centering
  \includegraphics[width=\linewidth]{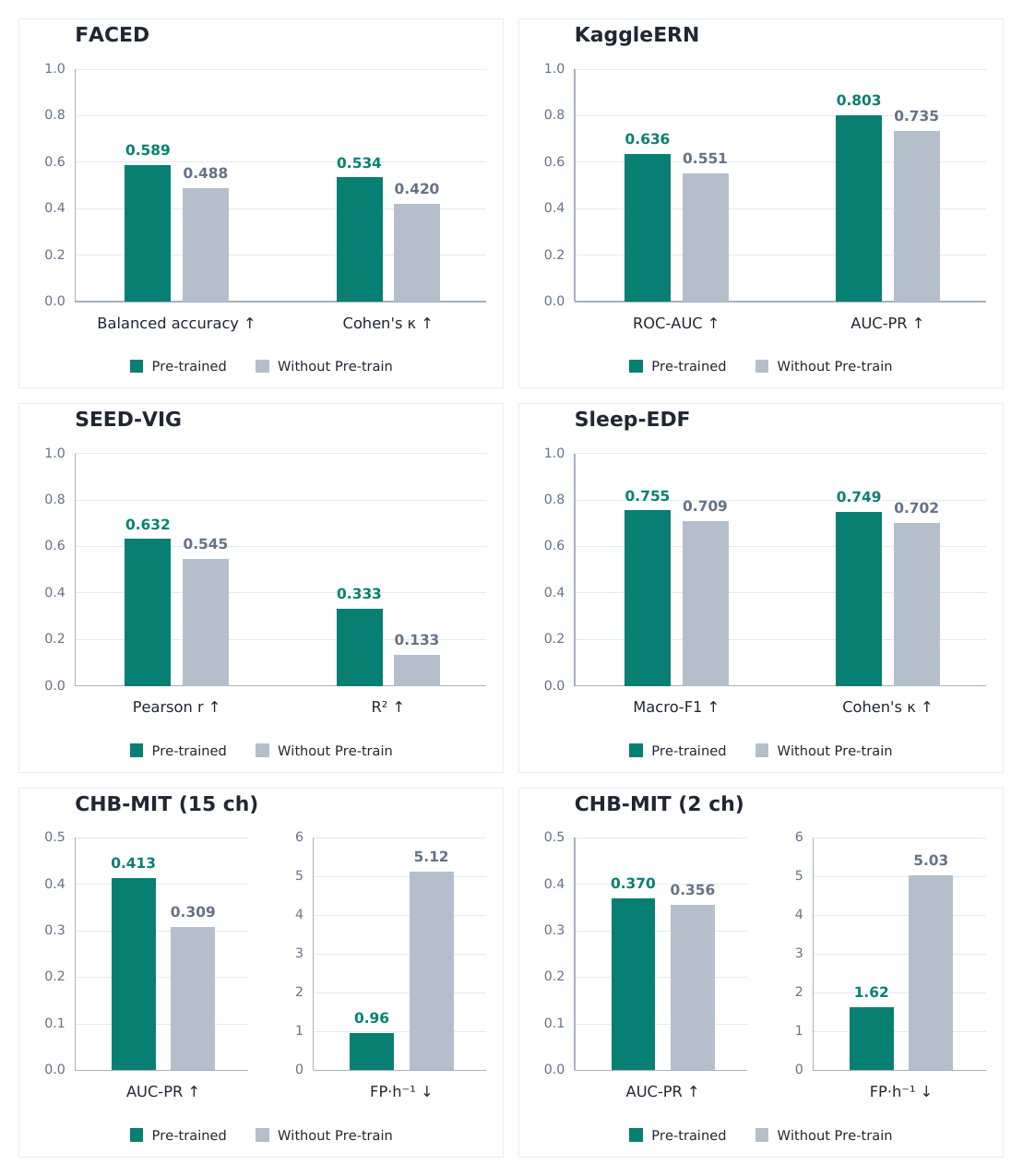}
  \caption{\textbf{Effect of pre-training.} Downstream performance of NeurDuo (Small) with and without pre-training. Arrows indicate whether higher or lower values are better.}
  \label{fig:no_pretrain}
\end{figure}

In this experiment, we fine-tune NeurDuo (Small) on every downstream task from random initialisation, keeping the architecture, shared readout, context head, data splits and optimisation recipe identical to the pre-trained runs; only the initial backbone weights differ. As shown in Figure~\ref{fig:no_pretrain}, pre-training improves performance on all five tasks.

\subsection{Two readings of the sleep-staging result}
\label{app:seqdiag}
Table~\ref{tab:long_sequence_results} reports one row per model on Sleep-EDF and Table~\ref{tab:state_carrying} a difference of two means. Both compress something so we give the per-class and the per-fold view here.

\paragraph{Where the macro-F1 gap sits.} Macro-F1 averages five per-class scores, and on this task those five are not equally hard. Table~\ref{tab:perclass} splits the results of Table~\ref{tab:long_sequence_results} by class. Our deficit against LaBraM is not spread across the table: we lead on N1, the class every model finds hardest and where the whole field sits between $0.41$ and $0.48$, we are best on REM, and we are within $0.010$ of LaBraM on W. The entire $0.011$ macro-F1 gap comes from N3, where we trail by $0.060$; on the other four classes we are collectively ahead of LaBraM. N3 is defined by absolute delta amplitude sustained over an epoch, which is the one thing in this task a single $30$\,s window already contains. The rows below the rule show what carrying state does to exactly that class: with the state reset the N3 gap widens to $0.096$. Carrying state raises N3 F1 from $0.703$ to $0.739$, closing over a third of that gap, the largest per-class movement anywhere in the table and the one that contributes most to the macro-F1 improvement of Table~\ref{tab:state_carrying}.

\paragraph{Why the dispersion in Table~\ref{tab:state_carrying} does not bound the effect.} The carried setting improves macro-F1 by $0.010$ while the fold-to-fold standard deviation reaches $0.014$, so read as two independent measurements the intervals overlap and the effect looks unresolved. They are not independent. Both settings run on the same five subject-disjoint folds with the same seed, so the difference can be taken within each fold, and the fold-to-fold variation in how separable a group of subjects happens to be cancels. Table~\ref{tab:paired} does that. All fifteen differences are positive, and the paired $t$ statistics for accuracy and $\kappa$ exceed the two-sided $5\%$ critical value at four degrees of freedom, $2.776$. The corresponding statistic for macro-F1 remains below this threshold. We report the aggregate in the main text and this decomposition here rather than significance stars in the table itself, since with five folds a $t$ test is a description of consistency more than a test.

\begin{table}[!t]
\centering
\caption{\textbf{Per-class F1 on Sleep-EDF}, five-fold means. Classes are the five AASM stages, W for wake and N1 to N3 for non-REM depth. \textbf{Bold} marks the best value in a column among the seven models above the rule, each in the configuration Table~\ref{tab:long_sequence_results} reports. Below the rule are that same model with its state reset at every unit boundary and the difference between the two, neither of which is part of that comparison. $^\dagger$Domain overlap on Sleep-EDF, see Appendix~\ref{app:overlap}.}
\label{tab:perclass}
\begingroup
\fontsize{7}{9}\selectfont
\setlength{\tabcolsep}{2.5pt}
\renewcommand{\arraystretch}{1.08}
\begin{tabular}{@{}lccccc@{}}
\toprule
\textbf{Model}
& \textbf{W}
& \textbf{N1}
& \textbf{N2}
& \textbf{N3}
& \textbf{REM} \\
\midrule
BIOT$^\dagger$
& $0.924$ & $0.463$ & $0.828$ & $0.676$ & $0.764$ \\
CBraMod
& $0.914$ & $0.458$ & $0.845$ & $\mathbf{0.805}$ & $0.775$ \\
LaBraM
& $0.928$ & $0.478$ & $\mathbf{0.851}$ & $0.799$ & $0.773$ \\
EEGPT
& $0.922$ & $0.455$ & $0.829$ & $0.766$ & $0.725$ \\
ST-EEGFormer-S
& $0.923$ & $0.413$ & $0.826$ & $0.672$ & $0.740$ \\
REVE-Base
& $\mathbf{0.930}$ & $0.478$ & $0.835$ & $0.680$ & $0.784$ \\
NeurDuo (Small)
& $0.918$ & $\mathbf{0.484}$ & $0.841$ & $0.739$ & $\mathbf{0.794}$ \\
\midrule
NeurDuo (Small), state reset
& $0.919$ & $0.490$ & $0.829$ & $0.703$ & $0.783$ \\
$\Delta$ from carrying state
& $-0.001$ & $-0.006$ & $+0.012$ & $+0.036$ & $+0.011$ \\
\bottomrule
\end{tabular}
\endgroup
\end{table}

\begin{table}[!htbp]
\centering
\caption{\textbf{Carried minus reset on Sleep-EDF, fold by fold.} Same folds, same seed, same readout and context head; only the backbone state differs. $t$ is the paired statistic over the five differences, for which the two-sided $5\%$ critical value at four degrees of freedom is $2.776$.}
\label{tab:paired}
\begingroup
\fontsize{7}{9}\selectfont
\setlength{\tabcolsep}{2.5pt}
\renewcommand{\arraystretch}{1.08}
\begin{tabular}{@{}lcccccccc@{}}
\toprule
\textbf{Metric}
& \textbf{Fold 1}
& \textbf{Fold 2}
& \textbf{Fold 3}
& \textbf{Fold 4}
& \textbf{Fold 5}
& \textbf{Mean}
& $\mathbf{t}$
& \textbf{Positive} \\
\midrule
Macro-F1
& $+0.0277$ & $+0.0159$ & $+0.0039$ & $+0.0002$ & $+0.0037$
& $+0.0103$ & $+2.02$ & $5/5$ \\
Accuracy
& $+0.0089$ & $+0.0146$ & $+0.0044$ & $+0.0083$ & $+0.0043$
& $+0.0081$ & $+4.27$ & $5/5$ \\
Cohen's $\kappa$
& $+0.0141$ & $+0.0189$ & $+0.0051$ & $+0.0107$ & $+0.0040$
& $+0.0106$ & $+3.80$ & $5/5$ \\
\bottomrule
\end{tabular}
\endgroup
\end{table}

\subsection{Two streams in operation}
\label{app:streams}
The fast and slow streams of \S\ref{sec:backbone} are given different update rates by construction rather than by training. Two measurements show what that produces, one read from the weights alone and one from a forward pass on real signal. Neither can be produced for a window encoder, which has no state whose timescale could be measured.

\begin{figure}[!t]
  \centering
  \includegraphics[width=\linewidth]{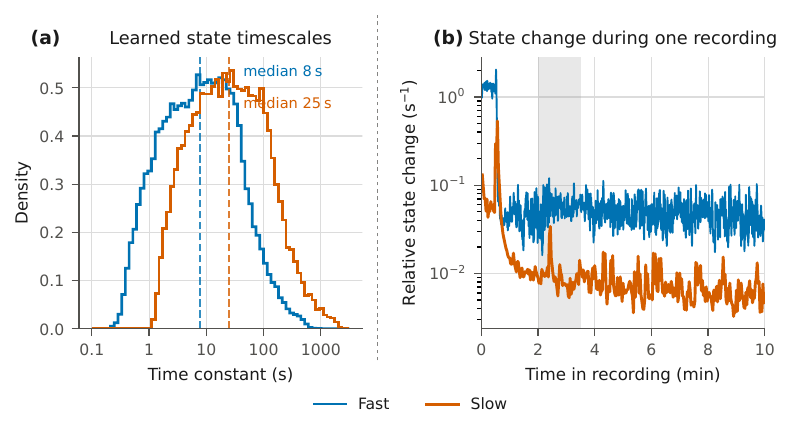}
  \caption{\textbf{The two streams run at different rates.} \emph{(a)} Nominal time constant $\tau = 1/(\Delta|A|)$, with $\Delta$ taken from the trained bias alone, of every state dimension of the pre-trained Small model, converted to seconds at each stream's own step; dashed lines are medians. \emph{(b)} Relative state change per second on one ten-minute CHB-MIT test sequence, averaged over the $15$ channels; the shaded band marks its three seizure units.}
  \label{fig:state}
\end{figure}

\paragraph{Timescales in the trained weights.} Figure~\ref{fig:state}(a) asks how fast individual state dimensions decay. In a selective state-space block the state decays by $\exp(\Delta A)$ at each step, so a dimension has a time constant of $\tau=1/(\Delta|A|)$ steps; $\Delta$ is input-dependent, and we report the nominal value $\mathrm{softplus}(b_\Delta)$ given by the trained bias alone, which sets the timescale before any input modulates it. We read $A$ and $b_\Delta$ off the weights of the Small model, take every combination of block, inner dimension and state dimension, and convert to seconds using each stream's own step, $0.5$\,s for the fast stream and $2$\,s for the slow one. Measured that way the slow stream is the longer-lived of the two, median $24.9$ against $7.7$\,s, and it reaches $35$ minutes where the fast stream stops at $14$. The separation in seconds largely reflects the update rate rather than the decays the optimiser found: in units of each stream's own step the two distributions largely overlap, medians $15.5$ and $12.4$ steps, so the $3.2\times$ separation in seconds follows from those medians together with the $N=4$ ratio between how often the two are advanced. That is the mechanism of \S\ref{sec:backbone} behaving as specified. It also locates what makes the slow stream slow: not mainly a decay the model had to learn, but how often it is written and what it is written with, summaries from the memory writer rather than the samples the fast stream sees.

\paragraph{Rates at inference.} Figure~\ref{fig:state}(b) asks whether the two streams also move at different rates when the model is running. On one ten-minute CHB-MIT test sequence we measure how much the representation each stream emits, $h_{t,c}$ for the fast stream and $s_{k,c}$ for the slow one, changes from one of its own steps to the next, relative to its own size, and divide by the step so that both are read per second of recording rather than per update. These are the two vectors Eq.~\ref{eq:readout} splices, not the internal SSM states of the Mamba blocks, so the figure shows the rate at which what the readout sees is rewritten. The fast representation changes by $0.0500$ per second at the median and the slow one by $0.0077$, a factor of six and a half, and the ordering holds at all but five of the $299$ points where the two can be compared. The difference is therefore not only in the weights: while the fast stream tracks the signal, the slow one barely moves. The fast stream takes about half a minute to settle from the zero state it starts a sequence in and the slow stream about a minute and a half, which is the cold start the half-sequence stride of Appendix~\ref{app:pretrain} is designed to make the model predict from, and which state carried across a unit boundary removes at evaluation time. The shaded band marks the sequence's three seizure units and is drawn for orientation only: on this one sequence neither stream shows a change there that stands out from its own baseline, and a single recording could not establish one if it did.

\section{Streaming Inference and Efficiency}
\label{app:streaminginference}
\subsection{Stateful Streaming Inference}
\label{app:streaming}
\begin{figure}[!t]
\centering
\includegraphics[width=\linewidth]{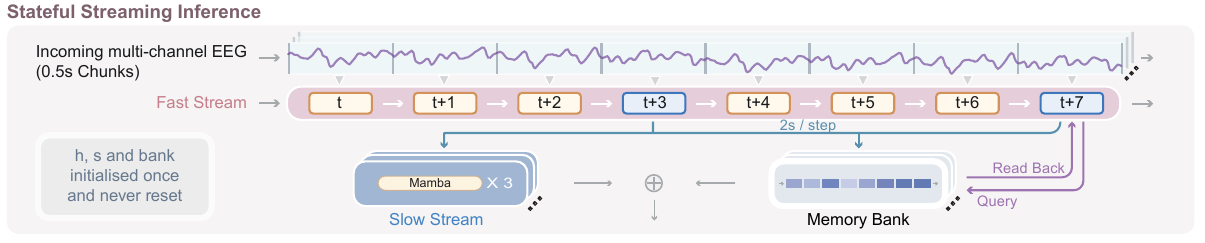}
\caption{\textbf{Stateful streaming inference.} Each $0.5$\,s chunk advances the fast stream by one step; every fourth chunk writes a memory token that advances the slow stream and enters the $16$\,s bank the fast state queries. Nothing is reset within a recording.}
\label{fig:streaming}
\end{figure}

Figure~\ref{fig:streaming} illustrates how NeurDuo runs on a continuous recording. For each channel, the fast and slow-stream Mamba states, the recent fast representations used by the memory writer and the memory bank are initialised once at the start of the recording and never reset. Each incoming $0.5$\,s chunk is encoded per channel and advances the fast stream by a single recurrent step, with channel attention mixing the channels at that step. Every four chunks ($2$\,s), the memory writer summarises the last eight fast representations of each channel into a memory token, which advances the slow stream by one step and is appended to the memory bank, evicting its oldest entry. At every chunk, the current fast representation queries the memory bank of its own channel, and the retrieved content is added to the slow representation before it is concatenated with the fast representation to form the output. No earlier chunk is ever encoded again, so the cost of each step does not depend on how much of the recording has already been processed (Appendix~\ref{app:cost}).

\subsection{The cost of long context}
\label{app:cost}
\S\ref{sec:backbone} gives NeurDuo a state that is bounded in size and advanced one chunk at a time. This appendix measures what that is worth against the alternative, which is to widen a window encoder until it spans the same history. All numbers are measurements on one A100 at float32. Latency depends on the shape of the input and not on its content, so the inputs are random tensors; what we borrow from the two sequence tasks is only their electrode configuration, two channels at $100$\,Hz and fifteen at $256$\,Hz, so that every model is given the same job on each.

\begin{table}[!htbp]
\centering
\caption{\textbf{Cost of conditioning on longer history.}
Latency (ms) to encode one window, with peak memory (GB) measured at
$1{,}800$\,s. NeurDuo latency is measured per $0.5$\,s chunk after
streaming the stated history.}
\label{tab:cost}
\begingroup
\fontsize{7}{9}\selectfont
\setlength{\tabcolsep}{2.5pt}
\renewcommand{\arraystretch}{1.08}

\begin{tabular}{@{}l*{12}{r}@{}}
\toprule
\multirow{2}{*}{\textbf{Model}}
& \multicolumn{6}{c}{\textbf{Sleep-EDF (2 ch)}}
& \multicolumn{6}{c}{\textbf{CHB-MIT (15 ch)}} \\
\cmidrule(lr){2-7}\cmidrule(lr){8-13}
& \textbf{30\,s}
& \textbf{300\,s}
& \textbf{600\,s}
& \textbf{1,800\,s}
& \textbf{3,600\,s}
& \textbf{GB}
& \textbf{30\,s}
& \textbf{300\,s}
& \textbf{600\,s}
& \textbf{1,800\,s}
& \textbf{3,600\,s}
& \textbf{GB} \\
\midrule

EEGPT-Large
& $4.0$ & $27.3$ & $49.8$ & $137.6$ & $274.0$ & $1.22$
& $8.5$ & $66.0$ & $128.2$ & $382.9$ & $765.5$ & $3.46$ \\

ST-EEGFormer-S
& $3.9$ & $40.5$ & $136.6$ & $1{,}221$ & $5{,}119$ & $0.97$
& $28.9$ & $1{,}934$ & $8{,}061$ & $74{,}302$ & --$^\ast$ & $5.64$ \\

BIOT
& $6.0$ & $23.3$ & $40.5$ & $106.9$ & $209.2$ & $0.89$
& $6.0$ & $23.3$ & $42.7$ & $107.0$ & $209.5$ & $0.98$ \\

LaBraM
& $9.9$ & $98.0$ & $196.4$ & $587.2$ & $1{,}174$ & $0.05$
& $10.0$ & $98.1$ & $196.2$ & $585.9$ & $1{,}177$ & $0.20$ \\

CBraMod
& $5.8$ & $5.9$ & $6.1$ & $15.1$ & $38.1$ & $0.16$
& $5.8$ & $9.2$ & $19.0$ & $93.6$ & $281.4$ & $0.99$ \\

REVE-Base
& $8.5$ & $12.5$ & $21.3$ & $90.9$ & $257.5$ & $0.40$
& $9.4$ & $115.3$ & $400.6$ & $3{,}653$ & $15{,}288$ & $1.16$ \\

\rowcolor{black!5}
\textbf{NeurDuo (Small)}
& $\mathbf{4.89}$ & $\mathbf{4.90}$ & $\mathbf{4.84}$
& $\mathbf{4.85}$ & $\mathbf{4.89}$ & $\mathbf{0.05}$
& $\mathbf{4.90}$ & $\mathbf{4.87}$ & $\mathbf{4.85}$
& $\mathbf{4.87}$ & $\mathbf{4.83}$ & $\mathbf{0.06}$ \\

\rowcolor{black!5}
\textbf{NeurDuo (Base)}
& $\mathbf{6.28}$ & $\mathbf{6.31}$ & $\mathbf{6.29}$
& $\mathbf{6.29}$ & $\mathbf{6.29}$ & $\mathbf{0.13}$
& $\mathbf{6.43}$ & $\mathbf{6.42}$ & $\mathbf{6.41}$
& $\mathbf{6.41}$ & $\mathbf{6.42}$ & $\mathbf{0.16}$ \\

\rowcolor{black!5}
\textbf{NeurDuo (Large)}
& $\mathbf{8.72}$ & $\mathbf{8.74}$ & $\mathbf{8.74}$
& $\mathbf{8.75}$ & $\mathbf{8.73}$ & $\mathbf{0.29}$
& $\mathbf{8.94}$ & $\mathbf{8.93}$ & $\mathbf{8.94}$
& $\mathbf{8.92}$ & $\mathbf{8.93}$ & $\mathbf{0.37}$ \\

\bottomrule
\end{tabular}
\endgroup

\vspace{2pt}
\parbox{\linewidth}{\footnotesize
$^\ast$Not measured: the $1{,}800$-s run already took $74.3$\,s.}
\end{table}

\paragraph{Widening a window against advancing a state.} Table~\ref{tab:cost} gives the measurements behind Figure~\ref{fig:cost}, on both montages and with peak memory. As released, four of the six baselines stop short of an hour, but in each case the limit is a fixed-size table or a constant fixed when the encoder is built, not the architecture, so we lift it without adding a parameter or changing a learned weight. BIOT's $1{,}000$-row and ST-EEGFormer's $512$-row position tables bound the input at $500$ and $64$ seconds respectively, but both are computed from a sinusoidal formula rather than trained, so we regenerate them at the required length and verify that the released rows are reproduced exactly. EEGPT places patches in time by rotary encoding and attends within a time patch, so no table bounds its input and only the expected patch count has to be set to that of the window. LaBraM's time table is learned and has $16$ rows, which cannot be extended, so the window is encoded as consecutive $15$\,s sub-windows. CBraMod and REVE run natively. Every baseline therefore takes every length we report, and what the table measures is the price of the alternative: our rows give the cost of one further $0.5$\,s chunk after the stated history has been streamed, which for a window encoder is the cost of the whole window, since it has to be encoded again to take in the new chunk. None of this affects any number in Tables~\ref{tab:long_sequence_results}, \ref{tab:context_ladder}, \ref{tab:ablations}, \ref{tab:state_carrying}, where no baseline is shown more than one $30$\,s unit and the shared context head integrates across units.

\paragraph{What the state costs.} Two quantities are easy to conflate and we separate them. The first is the state itself, written $\Sigma_t$: the convolution and SSM states of the fast blocks, the SSM states of the slow blocks, the queue of recent fast representations the writer pools over, and the $B$ slots of the bank. It is what has to survive from one chunk to the next, its size follows from the architecture rather than from the recording, and Table~\ref{tab:state} breaks it down for NeurDuo (Small). At $349$\,KB per channel it comes to $0.68$\,MB for the two channels of Sleep-EDF and $5.12$\,MB for the fifteen of CHB-MIT. The second quantity is the memory a step occupies while it runs, which additionally holds the parameters and the activations of the one chunk being processed: $0.047$ and $0.061$\,GB for Small on the two montages.

\begin{table}[!t]
\centering
\caption{\textbf{The persistent state}, per channel, for NeurDuo (Small). Sizes follow from the architecture constants of Table~\ref{tab:arch} and are the same at every point in a recording. The writer queue is a ring buffer of $16$ slots, of which each write pools the most recent $w=8$; an implementation that kept only those eight would save $8$\,KB per channel.}
\label{tab:state}
\begingroup
\fontsize{7}{9}\selectfont
\setlength{\tabcolsep}{2.5pt}
\renewcommand{\arraystretch}{1.08}
\begin{tabular}{@{}llrr@{}}
\toprule
\textbf{Component}
& \textbf{Shape}
& \textbf{Values}
& \textbf{KB (fp32)} \\
\midrule
Fast stream
& $6\times(512\times16\ \text{SSM}+512\times4\ \text{conv})$
& 61,440
& 240.0 \\
Slow stream
& $3\times(384\times16\ \text{SSM}+384\times3\ \text{conv})$
& 21,888
& 85.5 \\
Memory bank
& $8\times192$
& 1,536
& 6.0 \\
Writer queue
& $16\times256$
& 4,096
& 16.0 \\
Last output
& $256+192$
& 448
& 1.8 \\
\midrule
\textbf{Total per channel}
&
& \textbf{89,408}
& \textbf{349.2} \\
\bottomrule
\end{tabular}
\endgroup
\end{table}

\paragraph{Streaming an hour.} Table~\ref{tab:stream} follows both quantities along a stream, for NeurDuo (Small). Latency is the cost of one chunk averaged over a full write cycle of four chunks, three that only advance the fast stream and one that also writes a memory token and advances the slow stream, $4.5$ to $4.8$\,ms and $5.8$\,ms respectively, which average to the $4.9$\,ms per chunk the table reports; the cycle is timed $25$ times after $64$ warm-up chunks, with the state restored before each repetition so that measuring does not advance the stream. Peak memory is the high-water mark of allocated memory since streaming began, covering parameters, state and activations but not the CUDA context, so anything accumulated across the stream would make it climb. Over the six points the state does not change by a byte, peak memory holds to three decimal places, and latency varies by $1.5\%$, the spread of repeated timings on an idle device. A chunk carries $0.5$\,s of signal and takes $4.9$\,ms to absorb, so the model runs about a hundred times faster than the recording arrives, and almost as fast at fifteen channels as at two, because the channels advance as parallel rows rather than as a longer sequence; what scales with the montage is the state, linearly and by construction.

\begin{table}[!htbp]
\centering
\caption{\textbf{Streaming a recording from thirty seconds to an hour}, NeurDuo (Small). Each row advances the state to the stated amount of history and then measures the next chunks, averaged over a full write cycle. Nothing in the table depends on the history, which is the property \S\ref{sec:backbone} claims and this table exists to check.}
\label{tab:stream}
\begingroup
\fontsize{7}{9}\selectfont
\setlength{\tabcolsep}{2.5pt}
\renewcommand{\arraystretch}{1.08}
\begin{tabular}{@{}lrrrrcrrrr@{}}
\toprule
& \multicolumn{4}{c}{\textbf{Sleep-EDF montage, two channels}}
&
& \multicolumn{4}{c}{\textbf{CHB-MIT montage, fifteen channels}} \\
\cmidrule(lr){2-5}\cmidrule(lr){7-10}
\textbf{History}
& \textbf{ms}
& $\boldsymbol{\times}$ \textbf{real time}
& \textbf{GB peak}
& \textbf{MB state}
&
& \textbf{ms}
& $\boldsymbol{\times}$ \textbf{real time}
& \textbf{GB peak}
& \textbf{MB state} \\
\midrule
$30$\,s
& $4.89$ & $102\times$ & $0.047$ & $0.68$
&
& $4.90$ & $102\times$ & $0.061$ & $5.12$ \\
$1$\,min
& $4.89$ & $102\times$ & $0.047$ & $0.68$
&
& $4.89$ & $102\times$ & $0.061$ & $5.12$ \\
$5$\,min
& $4.90$ & $102\times$ & $0.047$ & $0.68$
&
& $4.87$ & $103\times$ & $0.061$ & $5.12$ \\
$10$\,min
& $4.84$ & $103\times$ & $0.047$ & $0.68$
&
& $4.85$ & $103\times$ & $0.061$ & $5.12$ \\
$30$\,min
& $4.85$ & $103\times$ & $0.047$ & $0.68$
&
& $4.87$ & $103\times$ & $0.061$ & $5.12$ \\
$60$\,min
& $4.89$ & $102\times$ & $0.047$ & $0.68$
&
& $4.83$ & $104\times$ & $0.061$ & $5.12$ \\
\bottomrule
\end{tabular}
\endgroup
\end{table}

\section{Discussion}
\label{app:discussion}
\paragraph{Limitations.} Three are worth stating plainly. First, these results suggest that increasing model capacity alone yields limited and task-dependent gains under the current pre-training setup, motivating larger pre-training corpora. We pre-train on $3{,}955$ hours, an order of magnitude less than the largest corpus among our baselines, and Figure~\ref{fig:pretrain-curves} shows what that costs: the Large model reaches its minimum $51{,}000$ updates before the schedule ends and then rises, while Small and Base are still descending when it ends, and their minima lie within $0.03$ nats of one another. The downstream tables say the same thing from the other side, since Large adds nothing over Base on FACED and loses to it on SEED-VIG. Second, the advantage is uneven across the two sequence tasks, and the pattern is informative rather than incidental. On CHB-MIT, which supplies fifteen electrodes, scaling from Small to Large lifts AUC-PR from $0.413$ to $0.471$; on Sleep-EDF, which supplies two derivations, the three sizes are separated by $0.009$ macro-F1 and we are not first, trailing LaBraM by $0.011$. Appendix~\ref{app:seqdiag} locates that gap in a single class rather than across the five, which is consistent with a channel-parallel frontend having little to separate when the montage is two channels wide. Third, the benefit of context is real but bounded, and the bound is a property of the task. Sleep staging benefits overall from longer context, with smaller gains beyond ten minutes. Seizure detection does not: with the full montage the optimum sits between ten and thirty minutes, and the two metrics do not agree on where inside that range it falls. Longer context is therefore not a direction that pays indefinitely, and the useful scale has to be found per task rather than assumed.

\paragraph{Outlook.} Three directions follow from those limits. The first is simply more continuous signal: the architecture is indifferent to montage and to recording length, so the obstacle to a larger model is the supply of long recordings rather than anything in the design, and the curves above say that is where the next gain is. The second is to make consolidation and retrieval adaptive. The writer currently fires on a fixed schedule, every $N=4$ steps, and the bank holds a fixed span of $16$ seconds; both are constants chosen once and held across the family, whereas what should be written and how far back should be read are plausibly functions of the signal. A writer that fires on change rather than on a clock, and a bank whose span adapts to the timescale a task needs, are the natural next version of the mechanism this paper fixes by construction. The third is evaluation in the setting the model was built for. Every result here is offline, scored after the fact on complete recordings, yet the state is bounded and causal precisely so that inference can run alongside an ongoing session.

\end{document}

%% file: math_commands.tex
\usepackage{amsmath,amsfonts,bm}

\def\eqref#1{equation~\ref{#1}}

\def\1{\bm{1}}

\DeclareMathAlphabet{\mathsfit}{\encodingdefault}{\sfdefault}{m}{sl}
\SetMathAlphabet{\mathsfit}{bold}{\encodingdefault}{\sfdefault}{bx}{n}

